\documentclass{article}

\usepackage{geometry}

\usepackage{subfigure} 

\usepackage{authblk}

\usepackage{amsmath,amsthm}
\usepackage{amssymb}
\usepackage{epsfig}
\usepackage{hyperref}
\usepackage{color}

\usepackage{enumitem}

\newcommand{\beq}{\begin{equation}}
\newcommand{\eeq}{\end{equation}}

\newcommand{\bdisp}{\begin{displaymath}}
\newcommand{\edisp}{\end{displaymath}}

\newcommand{\beqarr}{\begin{eqnarray}}
\newcommand{\eeqarr}{\end{eqnarray}}

\newcommand{\bmlt}{\begin{multline}}
\newcommand{\emlt}{\end{multline}}

\newcommand{\beqarrn}{\begin{eqnarray*}}
\newcommand{\eeqarrn}{\end{eqnarray*}}

\newcommand{\benum}{\begin{enumerate}}
\newcommand{\eenum}{\end{enumerate}}

\newcommand{\bit}{\begin{itemize}}
\newcommand{\eit}{\end{itemize}}

\newcommand{\bc}{\begin{center}}
\newcommand{\ec}{\end{center}}

\newcommand{\bdes}{\begin{description}}
\newcommand{\edes}{\end{description}}

\newcommand{\bfig}{\begin{figure}}
\newcommand{\efig}{\end{figure}}

\newcommand{\bemq}{\begin{quote} \begin{em}}
\newcommand{\eemq}{\end{em} \end{quote}}

\newcommand{\bmp}{\begin{minipage}}
\newcommand{\emp}{\end{minipage}}

\newcommand{\eqn}[1]{(\ref{#1})}

\newcommand{\indic}[1]{\mathbf{1}{\left\{{#1}\right\}}}

\newcommand{\define}{\triangleq}

\newcommand{\ie}{{i.e.}}

\newcommand{\eg}{{ e.g.}}

\newcommand{\st}{\text{ s.t. }}

\newcommand{\bsp}{\begin{slide*}}
\newcommand{\esp}{\end{slide*}}
\newcommand{\bsl}{\begin{slide}}
\newcommand{\esl}{\end{slide}}

\newcommand{\virag}[1] {{[\bf virag: #1]}}
\newcommand{\revised}[1]{{#1}}
\renewcommand{\virag}[1]{}

\newtheorem{definition}{Definition}
\newtheorem{theorem}{Theorem}
\newtheorem{lemma}{Lemma}

\newtheorem{proposition}{Proposition}

\title{Adaptive Matching for Expert Systems with Uncertain Task Types}
\author[1]{Virag Shah}
 \author[2]{Lennart Gulikers}
 \author[2]{Laurent Massouli\'e}
 \author[3]{ Milan Vojnovi\'c}
 \affil[1]{Stanford University}
\affil[2]{Microsoft Research-INRIA Joint Centre}
\affil[3]{London School of Economics}
\begin{document}

\maketitle

\begin{abstract}
A matching in a two-sided market often incurs an externality: a matched resource may become unavailable to the other side of the market, at least for a while. This is especially an issue in online platforms involving human experts as the expert resources are often scarce. The efficient utilization of experts in these platforms is made challenging by the fact that the information available about the parties involved is usually limited. 

To address this challenge, we develop a model of a task-expert matching system where a task is matched to an expert using not only the prior information about the task but also the feedback obtained from the past matches. In our model the tasks arrive online while the experts are fixed and constrained by a finite service capacity. For this model, we characterize the maximum task resolution throughput a platform can achieve. We show that the natural greedy approaches where each expert is assigned a task most suitable to her skill is suboptimal, as it does not internalize the above externality. We develop a throughput optimal backpressure algorithm which does so by accounting for the `congestion' among different task types. Finally, we validate our model and confirm our theoretical findings with data-driven simulations via logs of Math.StackExchange, a StackOverflow forum dedicated to mathematics.
\end{abstract}

%



\section{Introduction}
\label{sec:intro}

Online platforms that enable matches between trading partners in two-sided markets have recently blossomed in many areas: 
LinkedIn and Upwork facilitate matches between employers and employees; Uber allows matches between passengers and car drivers; Airbnb and Booking.com connect travelers and housing facilities; Quora and Stack Exchange facilitate matches between questions and  either answers, or experts able to provide them. 

\revised{These platforms often propose matches based on imperfect knowledge of the characteristics of the two parties to be matched. Such uncertainty may result into inferior matches and may incur negative externalities of the following kind:  If a constrained resource is matched sub-optimally then it becomes unavailable to a more suitable match for a while. For example, in online labour platforms and Q\&A platforms if an expert is matched to a task which does not meet her expertise then the tasks which meet her expertise may suffer. Similarly, in hospitality platforms an economical accommodation becomes unavailable to a financially constrained customer if it is matched to a flexible customer. }



\revised{This naturally leads to the following questions: 
\begin{itemize}
\item How to quantify the loss in efficiency resulting from such uncertainty?
\item Which matching recommendation algorithms can lead to the most efficient platform operation in presence of such uncertainty? 
\end{itemize}
A natural measure of efficiency is the throughput that the platform achieves, i.e. the rate of successful matches it allows. 

In this paper, we progress towards answering these questions as follows. In what follows, we will anchor our discussion to task-expert systems but the insights developed are more generally applicable. }

First, we propose a simple model of such platforms, which features a static collection of servers, or experts on the one hand, and a continuous stream of arrivals of tasks, or jobs, on the other hand. In our model, the platform's operation consists of servers iteratively attempting to solve tasks. After being processed by some server, a task leaves the system if solved; otherwise it remains till successfully treated by some server. 
To model uncertainty about task types, we assume that for each incoming task we are given the prior distribution of this task's ``true type''. Servers' abilities are then represented via the probability that each server has to solve a task of given type after one attempt at it. 

In a Q\&A platform scenario, tasks are questions, and servers are experts; a server processing a task corresponds to an expert providing an answer to a question. A task being solved corresponds to an answer being accepted. In an online labour platform, tasks could be job offers, and a server may be a pool of workers with similar abilities. A server processing a task then corresponds to a worker being interviewed for a job, and the task is solved if the interview leads to a hire. We could also consider the dual interpretation when the labour market is constrained by workers rather than job offers. Then a task is a worker seeking work, while a server is a pool of employers looking for hires. 

An important feature of our model consists in the fact that when a task's processing does not lead to success, it does however affect uncertainty about the task's type. Indeed, the a posteriori distribution of the task's type after a failed attempt on it by some server differs from its prior distribution. For instance in a Q\&A scenario, a question which an expert in Calculus failed to answer either is not about Calculus, or is very hard. \revised{Further, the feedback from the expert may reveal some information about the task's type.}

For our model, we then determine necessary and sufficient conditions for an incoming stream of task arrivals to be manageable by the servers, or in other words, determine achievable throughputs of the system. In the process we introduce candidate policies, in particular the greedy policy according to which a server choses to serve tasks for which its chance of success is highest. This scheduling strategy is both easy to implement and is based on a natural motivation. Surprisingly perhaps, we show that it is not optimal in the throughput it can handle. In contrast, we introduce a so-called backpressure policy inspired from the wireless networking literature \cite{TaE92}, which we prove to be throughput-optimal.

We summarize contributions of this paper as follows:
\begin{itemize}[noitemsep,topsep=0pt]
\item We propose a new model of a generic task-expert system that allows for uncertainty of task types, heterogeneity of skills, and recurring attempts of experts in solving tasks. 
\item We provide a full characterization of the stability region, or sustainable throughputs, of the task-expert system under consideration. We establish that a particular backpressure policy is throughput-optimal, in the sense that it supports maximum task arrival rate under which the system is stable.  
\item We show that there exist instances of task-expert systems under which simple matching policies such as a natural greedy policy and a random policy can only support a much smaller maximum task arrival rate, than the backpressure policy.
\item We report the results of empirical analysis of the popular Math.StackExchange Q\&A platform which establish heterogeneity of skills of experts, with experts knowledgeable across different types of tasks and others specialized in particular types of tasks. We also show numerical evaluation results that confirm the benefits of the backpressure policy on greedy and random matchmaking policies.
\end{itemize} 

The remainder of the paper is structured as follows. In Section~\ref{sec:system} we present our system model. In Section~\ref{sec:stability} we present the throughput optimal algorithm as well as the characterization of task arrival rates that can be supported by the system. In Section~\ref{sec:strawman}, we present a case study where we compare performance of our algorithm with other baseline algorithms.  In Section~\ref{sec:exp}, we present our experimental results. In Section~\ref{sec:GeneralFeedback} we generalize our results to arbitrary feedback structure. Related work is discussed in Section~\ref{sec:related}. We conclude in Section~\ref{sec:conc}. Proofs of the results are provided in Section~\ref{sec:proofs}.


\section{Problem Setting}
\label{sec:system}
Let $C = \{c_1,\ldots, c_k\}$ be the set of task types. Each task in the system is of a particular type in $C$. Let $S = \{s_1,\ldots,s_m\}$ be the set of servers (or experts) present in the system. When a server $s\in S$ attempts to resolve a task of type $c \in C$, the outcome is $1$ (a success) with probability $p_{s,c}$ and it is $0$ (a failure) with probability $1-p_{s,c}$. Upon success we say that the task is resolved. In the context of online hiring platform, this is equivalent to successful hiring of an employee for a job. In the context of Q\&A platform, this is equivalent to an answer by an expert being accepted by the asker of the question. 

We consider a Bayesian setting where we have a prior distribution $z=(z_c)_{c\in C}\in \mathcal C$ for a task's type, where $\mathcal C$ is the set of all distributions. Note, different tasks may have different prior distributions. Clearly, if server $s$ processes a task with prior distribution $z$ then the probability that it fails is given by
\begin{equation}
\psi_s(z) = \sum_{c\in C} z_c (1-p_{s,c}).
\end{equation}

Further, upon failure, the posterior distribution of task's type is given by 
\begin{equation}
\phi_s(z) = \left( \frac{z_c(1-p_{s,c})}{\psi_s(z)}\right)_{c \in C}.
\end{equation}

 Note that the posterior distribution of a task's type upon failure by a subset of servers does not depend on the sequence in which these servers resolve the task, \ie, for each $s,s' \in S$ we have $\phi_s \circ \phi_{s'} = \phi_{s'} \circ \phi_s$.  At any point in time a task is associated with a `mixed-type' which is defined as the posterior distribution of its type given the past attempts. 
 
We allow a task to be attempted sequentially by multiple servers until it is resolved. We would like to resolve the tasks as quickly as possible. \revised{The matching algorithm may use the past feedback from the servers. In the setting described above the feedback is binary, namely, in the form of success and failure. More generally, the servers may provide a more detailed feedback. Although in several cases such a feedback is not reliable and often biased, \eg, see \cite{DDP15}. For now, we will stick with the binary feedback structure. We will generalize our results to an arbitrary feedback structure in Section~\ref{sec:GeneralFeedback}.}

\subsection{Single Task Scenario}\label{sec:OneTaskCase} 

Before considering the setting of online task arrivals, for ease of exposition we first consider a toy scenario with single task for which greedy algorithms are known to be approximately optimal. Suppose that time $t\in \mathbb Z_+$ is discrete.  A task arrives at time $t=0$.  Let the prior distribution of its type upon arrival (equivalently, its mixed-type at time $t=0$) be $z$. At a time, only one server attempts to resolve a task. Consider the problem of designing a sequence of servers $(s(t): 0 \le t \le \tau)$ such that the probability that the task is resolved within a fixed time $\tau$ is maximized. Let $z(0) = z$, and for each $t\ge 1$ let $z(t) = \phi_{s(t-1)}(z(t-1))$, \ie, $z(t)$ is the mixed-type of the task at time $t$ given that it was not resolved upon previous attempts. Then the probability that the task is resolved by time $\tau$ is given as $g\big((s(t): 0 \le t \le \tau)\big) = 1 - \prod_{t=0}^{\tau} \psi_{s(t)}(z(t))$.

Contrast this with the Bayesian active learning setting in \cite{GKR10,JCK14} where the goal is to reduce uncertainty in true hypothesis via outcome from several experiments.  Using a diminishing returns property called adaptive submodularity the authors in \cite{GKR10} obtain a policy which is competitive with the optimal. In our setting, $g$ is a submodular function. Thus a greedy policy where $s(t)$ for each $t$ is chosen to be from $\arg\min_s \psi_s(z(t))$ is $1 - 1/e$-competitive, see \cite{NeW88}.  


Further, in this paper we add an extra dimension to the problem which was not considered in the \cite{GKR10,JCK14}, namely, we consider the setting of online task arrivals where tasks of different mixed-types may compete for the servers resources before they leave upon being resolved. We design throughput optimal policies under such a setting.



\subsection{Online Task Arrivals}

We consider a continuous time setting, \ie, $t \in \mathbb R_+$. Tasks arrive at a rate of $\lambda$ per time unit on average. The mixed types of incoming tasks upon arrival are assumed i.i.d., taking values in a countable subset $\mathcal Z$ of $\mathcal C$. For each $z\in \mathcal Z$, let $\pi_z$ denote the probability that a new arrival is of mixed type $z$. Finally, the time for server $s \in S$ to complete an attempt on a task takes on average $1/\mu_s$ time units, and such attempt durations are i.i.d.. All involved sources of randomness are independent. 

We assume  that $\mathcal Z$ is closed under $\phi_s(\cdot)$, \ie, for each $z \in \mathcal Z$,  $\phi_s(z) \in \mathcal Z$. This loses no generality, as the closure of a countable set with respect to a finite number of maps $\phi_s$ remains countable.

We assume that a given task may be inspected several times by a given server and assume that the outcomes success / failure are independent at each inspection. This can be justified if a label $s$ in fact represents a collection of experts with similar abilities, in which case multiple processings by $s$ correspond to processing by distinct individual experts. 

For such a setting we would like to minimize the expected sojourn time of a typical task, \ie, the expected time between the arrival and the resolving of a typical task. Recall that the success probabilities $p_{s,c}$ are assumed to be arbitrary. Under such a heterogeneous setting minimizing expected sojourn time is a hard problem. In fact, this is true even when there is no uncertainty in task types. As a proxy to sojourn time optimal policies, we will strive for throughput optimal policies. In particular, we will characterize the arrival rates $\lambda$ for which the system can be {\em stabilized}, i.e. for which there exists a scheduling policy which induces a time-stationary regime of the system's behavior. Indeed for a stable system the long term task resolution rate coincides with the task arrival rate $\lambda$, and thus throughput-optimal policies must make the system stable whenever this is possible. Note that for an unstable system the number of outstanding tasks accumulate over time and the expected sojourn time tends to infinity. 

Finally, for simplicity we assume more specifically that the tasks arrive at the instants of a Poisson process with intensity $\lambda$, and that the time for server $s$ to complete an attempt at a task follows an Exponential distribution with parameter $\mu_s$. These are continuous time analog of i.i.d.\ arrivals and independent departures per time slot in discrete time setting. These assumptions will imply that the system state at any given time $t$ can be represented as a Markov process, which simplifies throughput analysis. The system throughput is often insensitive to such statistical assumptions on arrival and service times, \eg, see \cite{HJX05}.

We close the section with some additional assumptions and notations which will aid our analysis.

For each time $t$ let $N_z(t)$ represent the number of tasks of mixed-type $z$ present in the system and $N(t)=(N_z(t))_{z\in \mathcal Z}$. We also let $z(s,t)$ denote the mixed type of the task that server $s$ works on at time $t$. For strategies such that  the servers select which task to handle based uniquely on the vector $N(t)$, the process $(N(t))_{t\ge 0}$ forms a continuous-time Markov chain (CTMC) (\cite{Bre13,Kum14}). The policies considered in this paper are studied by analyzing the associated CTMC. 

We allow a task to be assigned to multiple experts at a given time. Further, we allow both preemptive as well as non-preemptive policies. Recall, in a preemptive policy an expert may drop a task under service if a task of a new mixed-type becomes available, whereas in a non-preemptive policy an expert must wait for his task to be serviced before taking up a new one. 






\section{Optimal Stability}
\label{sec:stability}

Main goal of this section is to provide necessary and sufficient conditions for stability of the system, and to provide explicit policies which stabilize the system when the sufficient conditions are satisfied. 

We obtain stability conditions via capacity constraints and flow conservation constraints which capture the flow of tasks from one type to another upon service by an expert. For instance, if $\nu_{s,z}$ represents the flow of tasks of mixed-type $z$ served by expert $s$, a fraction $1- \psi_s(z)$ of it leaves the system due to success while the rest gets converted into a flow of type $\phi_s(z)$. The total arrival rate of flow of mixed-type $z$, \ie, $ \lambda \pi_z + \sum_{s \in S, z' \in \phi^{-1}_s(z)} \nu_{s,z'} \psi_s(z')$, must match the total service rate, \ie, $\sum_{s\in S} \nu_{s,z}$. Further, the total flow service rate expert $s$, \ie, $\sum_{z \in \mathcal Z} \nu_{s,z}$,  must be less than its service capacity $\mu_s$. The following is the main result of this section. 

\begin{theorem}\label{thm:OptimalStability}
Suppose there exists $s$ such that  $\min_c p_{s,c} >0$. If there exist non-negative real numbers $\nu_{s,z}$ for each $s\in S$ and each $z \in \mathcal Z$, and positive real numbers $\delta_s$ for each $s\in S$ such that the following hold:
\begin{align}
\forall z \in \mathcal Z, \quad \lambda \pi_z + \sum_{s \in S, z' \in \phi^{-1}_s(z)} \nu_{s,z'} \psi_s(z') = & \sum_{s\in S} \nu_{s,z}, \label{eq:FlowConservation} \\
\forall s \in S, \quad \sum_{z \in \mathcal Z} \nu_{s,z}  + \delta_s \le  \mu_s, \label{eq:FlowCapacity}
\end{align}
then there exists a policy under which the system is stable. If there does not exist non-negative real numbers $\nu_{s,z}$, for $s\in S$, $z \in \mathcal Z$ and non-negative real numbers $\delta_s$ for $s\in S$ such that the above constraints hold, then the system cannot be stable.  
\end{theorem}

We use the condition of existence of an expert $s$ such that  $\min_c p_{s,c} >0$ only for a technical reason to simplify our proof. We believe that the result holds even when this condition is not true. 

\revised{One may envisage obtaining a throughput optimal static randomized policy from a solution to \eqn{eq:FlowConservation} and \eqn{eq:FlowCapacity} which, for example, maximizes the minimum $\delta_s$. It is not clear if this policy would result into a stable solution. Consider the following plausible scenario. While the total slack available at each server is finite, the total number of queues is infinite since we have one queue for each mixed-type. Depending on the system parameters, the optimal solution may assign a positive slack to each queue. Then, the infimum over the slacks at different queues would be zero. This would make the system unstable. 

To avoid this pitfall, we find a finite set of mixed-types $\mathcal{Y}$ such that the overall arrival rate into queues corresponding to mixed-types $\mathcal Z \backslash \mathcal Y$ is sufficiently small. 
We then group the infinite number of queues corresponding to $\mathcal Z \backslash \mathcal Y$ into a virtual queue. We thus obtain a system with finite number of queues which consists of the virtual queue and the queues corresponding to the mixed-types in $\mathcal{Y}$. 
For this system we use a dynamic policy, provided below, which is motivated by the literature on backpressure policies for constrained queueing systems, \eg, see \cite{TaE92,GNT06}. 

One may also envisage a static randomized policy obtained via a solution to a modified version of the constraints \eqn{eq:FlowConservation} and \eqn{eq:FlowCapacity} which would stabilize the above finite queueing system. Indeed, such a policy exists and we use its existence to show throughput optimality of our backpressure policy. Such a static policy, however, suffers from a severe practical limitation. By randomly selecting a queue for each server, the policy splits its capacity across several queues
In contrast, in our policy each server serves only one queue with a high backlog.  It is well known that pooling of a server's capacity, as against fragmenting its capacity across several queues, achieves better performance due to gains from statistical multiplexing. In fact, the performance improvement scales with the number of queues. 

Further, an agile backlog based dynamic policy may offer several practical advantages over solving a high-dimensional optimization problem in real systems where the parameters used may change over time. Thus, we believe it is natural to consider a backpressure approach over a static approach. 
}

We now describe the our dynamic policy which achieves optimal stability. We need some more notation to describe the policy. Consider a set $\mathcal Y \subset \mathcal Z$. 
Let $X(t)$ be the number of tasks in the system at time $t$ which have mixed-type $z \in \mathcal Z \backslash \mathcal Y$ or have had a mixed-type $z \in \mathcal Z \backslash \mathcal Y$ in the past. 
Further, for each $z \in \mathcal Y$ let $\tilde X_z(t)$ be the number of tasks with mixed-type $z$ which have had a mixed-type in $ \mathcal Z \backslash \mathcal Y$ in the past. Also, for convenience, for each $z \in \mathcal Z \backslash \mathcal Y$, let $\tilde X_z(t)$ be the number of tasks with mixed-type $z$, \ie, $ \tilde X_z(t) = N_z(t)$ for each $z \in \mathcal Z \backslash \mathcal Y$. Thus, we have $X(t) = \sum_z \tilde X_z(t)$. 

Finally,  for each $z \in \mathcal Y$ let $\tilde N_z(t)$ be the number of tasks of mixed-type $z$ which have not had a mixed-type in $ \mathcal Z \backslash \mathcal Y$ in the past. Thus, for each $z \in \mathcal Y$ we have $N_z(t) = \tilde X_z(t) + \tilde N_z(t)$. For the rest of this section we suppress the dependence on $t$ for brevity in notation.

\revised{Our policy operates in two modes, Random mode and Backpressure mode. During Random mode, each server is assigned a task from $X$ at random. During Backpressure mode, a server $s$ is assigned a task of a mixed type in $\mathcal Y$ with the highest `expected backlog', where the expected backlog at mixed-type $z$ accounts for the congestion at $z$ as well as at $\phi_s(z)$.  Further, it also accounts for the fact that with probability $(1-\psi_s(z))$ the task may get resolved and leave the system without seeing the congestion at $\phi_s(z)$. The decision regarding which mode to operate in is based on the relative congestions at $X$ and $\mathcal Y$.}

\begin{definition}[Backpressure($\mathcal Y$) policy]\label{def:BPy} 
For a given $\mathcal Y$, let $X$ and $(\tilde N_z)_{z\in \mathcal Y} $ be as defined above. 
For each $s\in S, z\in \mathcal Y$ let 
$$w_{s,z}(\tilde N,X) = \begin{cases} 
\tilde N_z  - \psi_s(z) \tilde N_{\phi_s(z)}, & \text{ if } \phi_s(z) \in\mathcal  Y \\
\tilde N_z  - \psi_s(z) X , & \text{ if } \phi_s(z) \in  \mathcal Z \backslash \mathcal Y
\end{cases}.$$
For a given $(\tilde N,X)$, let 
$$ B_s(\tilde N,X) = \arg\max_{z'\in \mathcal Y: \tilde N_{z'} > 0} w_{s,z}(\tilde N,X).$$
If $$\sum_s \mu_s \max_{z\in \mathcal Y: \tilde N_z > 0} w_{s,z}(\tilde N,X) \ge X  \min_{c\in C} \sum_s \mu_s  p_{s,c}  $$ then each expert is assigned a task in $\tilde N_z$ where $z\in B_s(\tilde N,X) \subset \mathcal Y$ with ties broken arbitrarily.
Else, each expert serves a task in $X$ chosen uniformly at random. 
\end{definition}

Note that, under Backpressure($\mathcal Y$) policy, $\left((\tilde N_z)_{z\in \mathcal Y},(\tilde X_z)_{z\in \mathcal Z}\right)$ is a CTMC. The following theorem establishes throughput optimality of the Backpressure($\mathcal Y$) policy.

\begin{theorem}\label{thm:BP}
Suppose there exists a server $s$ such that  $\min_c p_{s,c} >0$. If the sufficient conditions for stability as given in the statement of Theorem \ref{thm:OptimalStability} are satisfied, then there exists a finite subset $\mathcal Y$ of $\mathcal Z$ such that the policy Backpressure($\mathcal Y$) stabilizes the system. 
\end{theorem}

In particular, the Backpressure($\mathcal Z$) policy is optimally stable for Asymmetric($a$) system as defined in Definition \ref{DEF:2C2Sexample}.

To prove Theorem \ref{thm:BP}, we use Lyapunov-Forster theorem to show stability. We use the following Lyapunov function:
$$
L(\tilde N, \tilde X) = \sum_{z\in \mathcal Y} \tilde N_z^2 +  \left(\sum_{z\in \mathcal Z} \tilde X_z\right)^2 =  \sum_{z\in \mathcal Y} \tilde N_z^2 + X^2.
$$ 

As such, proving this result requires significantly different approach as compared to stability proofs via quadratic Lyapunov functions of classical constrained queueing networks with finite number of queues. In particular, the flow equations do not directly give a stabilizing static policy. In fact, there does not exist a static policy which stabilizes the system at all feasible loads. To avoid this pitfall, we find a finite set $\mathcal Y$ such that the overall arrival rate into $\mathcal Z \backslash \mathcal Y$ is small, and `pool' the slack capacity at the servers to serve the infinite number of queues in $\mathcal Z \backslash \mathcal Y$.

The stability part of Theorem~\ref{thm:OptimalStability} follows from Theorem~\ref{thm:BP}. For the converse statement in Theorem~\ref{thm:OptimalStability}, we use system ergodicity.

We now provide an alternative policy which achieves stability under a more restrictive condition that $p_{s,c}$ are bounded away from $0$ and $1$, but with the advantage that it does not rely on the precise numbers of jobs $N_z$ sharing the same mixed type $z$, but rather on `local averages'. As such it may remain optimally stable even when the distribution of mixed types of incoming jobs is no longer assumed to be discrete.

\begin{definition}[Backpressure($\epsilon$) policy]
Partition set $\mathcal C$ into finitely many subsets $A_i$, $i=1,\ldots,l$, such that each $A_i$ has diameter at most $\epsilon$, that is for all $z,z'\in A_i$ we have
$
|z-z'|=\sum_{c}|z_c-z'_c|\le \epsilon.
$
We then define $N(A_i):=\sum_{z\in A_i}N_z$, and the backpressure with respect to server $s$ of a given $z$ as 
$$
w_{s,z}(N):=N(A_i)-\psi_s(z)N(A_j),
$$
where $i$ and $j$ are such that $z\in A_i$ and $\phi_s(z)\in A_j$. Then, each expert is assigned a task with mixed-type in
$$ A_s(N) = \arg\max_{z\in \mathcal Z:  N_{z} > 0} w_{s,z}(N),$$
with ties broken uniformly at random
\end{definition}

We then have the following: 

\begin{theorem}\label{thm:AltBP}
Suppose that there exists $\alpha>0$ such that for each $s,c$ we have
\begin{equation}\label{superfluous}
p_{s,c}\in [\alpha,1-\alpha].
\end{equation}
Suppose further that the sufficient conditions for stability as given in the statement of Theorem \ref{thm:OptimalStability} are satisfied. Then, there exists an $\epsilon>0$ sufficiently small such that the Backpressure($\epsilon$) policy stabilizes the system.
\end{theorem}

For its proof, we use the Lyapunov function $L(N) = \sum_{i} N(A_i)^2$. Again, the proof involves a significantly different approach as compared to stability proofs for standard constrained queuing networks with finite number of queues.  In particular, we develop and use new flow equations which account for not only the sets $A_i$ associated with the mixed-types of the tasks but also the lengths of the history of the tasks. 

Unlike backpressure policy proposed in \cite{TaE92} under a different setting, which was agnostic to system arrival rates, a set $\mathcal Y$ (or the $\epsilon$) such that the policy Backpressure($\mathcal Y$) (or policy Backpressure($\epsilon$)) stabilizes the system may depend on the value of $\lambda$. While the policies as stated may be complex to implement, it allows us to develop implementable heuristics which significantly outperform greedy policy. We demonstrate this in Section~\ref{sec:exp}.

\section{Asymmetric($a$) Systems: A Case Study}
\label{sec:strawman}

\revised{In this section we study a class of task-expert systems, namely Asymmetric($a$) systems, defined below. These systems resemble the $N$-system considered in the literature of queueing systems where the tasks types are assumed to be known, see \cite{Har98,BeW01,TeD10}. In particular, we study the loss in throughput due to uncertainty in task type, and also compare the performance of the optimal algorithm with some baseline policies, namely the Random policy and the Greedy policies.}

\begin{definition}[Asymmetric($a$) System] \label{DEF:2C2Sexample}
Fix $0<a<1$. In the Asymmetric($a$) system there are two task types $C = \{c_1,c_2\}$ and two experts $S = \{s_1,s_2\}$. Each arrival is equally likely to be of both types, \ie, $\pi_{z'} = 1$ where $z'$ satisfies $z'_c = 1/2$ for each $c\in C$, and $\pi_z = 0$ if $z\neq z'$.  Both experts provide responses at unit rate, \ie, $\mu_s=1$ for each $s$. Further, for class $c_1$ we have $p_{s,c_1} = 1$ for each $s\in S$, and for class $c_2$ we have  $p_{s_1,c_2} = a $, and $p_{s_2,c_2} = 0$. 
\end{definition}

For the Asymmetric($a$) system, if a task of mixed-type $z'$ receives a failure from either of the experts then its mixed type becomes $z''$ where $z''_{c_1} = 0$ and $z''_{c_2} = 1$. Thus, it is sufficient to assume that $\mathcal Z = \{z',z''\}$ where $z'_c = \frac{1}{2}$ for each $c\in C$, and $z''_{c} = \indic{c=c_2}$, where $\indic{A} = 1$ if $A$ is true and $0$ otherwise.   
 Further, it is easy to check that $\psi_{s_1}(z') = (1-a)/2$, $\psi_{s_1}(z'') = 1-a$, $\psi_{s_2}(z') = 1/2$, and $\psi_{s_2}(z'') = 1$.

\subsection{Loss in throughput due to uncertainty in task types} 
To understand the source of loss in throughput due to uncertainty, we first provide throughput of the Asymmetric($a$) system, and then compare it with an analogous system where true type is known. The following proposition uses the flow equations from Theorem~\ref{thm:OptimalStability}. Its detailed proof is provided in the Appendix. 

\begin{proposition}\label{prop:OptAsymmetricA}
There exists a policy which stabilizes the Asymmetric($a$) system if we have $\lambda < \min\left\{3a/(a+1), 2a \right\}$. Further, if $\lambda> \min\left\{3a/(a+1), 2a \right\}$ then no policy can stabilize the system.
\end{proposition}

Now suppose that the true type of each task is revealed upon arrival. Throughput of such systems can be computed using the well-known stability conditions for the flexible-server systems, \eg, see \cite{MaS04}; in particular, the throughput of the Asymmetric($a$) system if true types are known is equal to $2a$. 

Thus, for $a > 1/2$ there is a loss in efficiency of the system. In particular, for $a=1$ the throughput reduces by $25\%$. This can be reasoned as follows. For small values of $a$, the main system bottleneck is servicing of tasks of true type $c_2$ by server $s_1$ since this is the only server which can serve such tasks. Since server $s_2$ is not bottlenecked, in case of uncertain task types its extra capacity may be used to identify tasks of true type $c_2$. However, if the $a$ is large, then both the servers are bottlenecked and thus the wasteful use of $s_2$ in servicing tasks of true type $c_2$ results in loss of throughput. 

\subsection{Throughput under Random Policy:} Let us first define the Random policy and then provide an expression for the throughput.

\begin{definition}[Random Policy] In the Random policy each expert $s$ is assigned a task chosen uniformly at random from the pool of outstanding tasks.
\end{definition}

The following proposition provides throughput under Random policy for task expert systems in general, and the 
Asymmetric($a$) system in particular. Its proof is provided in the Appendix.

\begin{proposition}\label{prop:Random}
Under Random policy, a task-expert system is stable if and only if it satisfies the following:
$$ 
\lambda < \left( \sum_{c\in C} \frac{\sum_{z\in\mathcal Z} z_c \pi_z}{\sum_{s\in S} \mu_s p_{s,c}} \right)^{-1}.
$$
In particular, the Random policy stabilizes the Asymmetric($a$) system if and only if $\lambda < 4a/(2+a)$.
\end{proposition}

As expected, for the Asymmetric($a$) system the throughput under the Random policy is significantly lower than the optimal throughput. 

To prove the above result we use fluid limit approach developed in \cite{RyS92,Dai95,Mas07}. Let $X_c(t)$ be the number of tasks in the system of pure-type $c$.  Let $X(t) = (X_c(t))_c$. Roughly, given initial condition $X(0) = x$, we let $\lim_{k\to \infty} \frac{1}{k} X(0) = x$, and study $\lim_{k\to \infty}  \frac{1}{k} X(kt)$. We use the following Lyapunov function:
$$ L(X) = \sum_c X_c \log \left(\frac{X_c}{\gamma_c\sum_{c'}X_{c'}} \right),$$
where $\gamma_c \define \lambda  \frac{\sum_{z\in Z}  z_c \pi_z}{\sum_{s\in S} \mu_s p_{s,c}}$. 

\subsection{Throughput under Greedy Policies}
Following the discussion in Section \ref{sec:OneTaskCase}, a question arises: does a greedy approach work well even under the online setting? From throughput perspective, a natural greedy approach is one where each expert is assigned a task which best suits its skills. 

We will consider two greedy policies, a Preemptive Greedy policy and a Non-Preemptive Greedy policy. 
As we will see below, both the greedy policies are throughput suboptimal for the Asymmetric($a$) system. Intuitively, the reason for their suboptimality can be explained as follows. Note that for $a>0$ we have $ \psi_s(\tilde z') < \psi_s(\tilde z'') $ for each $s$. Thus, under the greedy policies each expert gives priority to the tasks of mixed-type $z'$. However, since only one expert can successfully serve the tasks of mixed-type $z''$, servicing of these tasks may become a bottleneck, especially for the small and moderate values of $a$. In such a scenario, a policy in which the expert $s_1$ would prioritize queue $z''$, especially when its length is relatively large, as done by the Backpressure policy, would achieve a better throughput. 

We first discuss the Preemptive Greedy policy and then the Non-Preemptive Greedy policy. 

\begin{definition}[Preemptive Greedy Policy] In the Preemptive Greedy policy, at each time an expert is assigned an outstanding task which maximizes its success probability, \ie, for each time $t$ such that $|N(t)| > 0$ we have
$$
z(s,t) \in A_s(N(t)) \triangleq \arg\min_{z:N_z(t) > 0} \psi_s(z),
$$
where ties are broken uniformly at random. 
\end{definition}

The following proposition provides throughput achieved by the Preemptive Greedy policy for the Asymmetric($a$) system. The main idea behind its throughput derivation can be intuitively explained as follows. Since both the servers give priority to the tasks of mixed-type $z'$ at each time, the corresponding queue acts as an M/M/1 queue with service rate $2$ and arrival rate $\lambda$. Since the fraction of time this queue is empty is $1 - \lambda/2$, the capacity available at server $s_2$ to server tasks of mixed-type $z''$ is $1 - \lambda/2$. Thus, maximum rate of service for tasks of mixed-type $z''$ is $a(1 - \lambda/2)$. Similarly, the arrival rate for tasks of mixed type $z''$ can be shown to be $\lambda(2-a)/4$. The stability condition follows by comparing these two. The formal proof of the proposition can be found in the Appendix.

\begin{proposition}\label{prop:PreemptiveGreedy}
The Preemptive Greedy policy stabilizes the Asymmetric($a$) system if and only if we have $\lambda < 4a/(2+a)$. 
\end{proposition}

A surprising implication of the above theorem is that, for $a = 1/2$, the Preemptive Greedy policy as well as the Random Policy achieve throughput equal to $4/5$. The optimal throughput is $25\%$ higher. This shows the importance of designing a matching policy which is cognizant of the system bottlenecks, such as the Backpressure policies designed in Section~\ref{sec:stability}. For the N-systems where the task types are known, it was first observed in \cite{Har98} that a greedy policy is suboptimal.

In the Preemptive Greedy policy, the process $(N(t))_{t\ge 0}$ is a CTMC. In particular, the order in which the tasks of a given mixed-type are served does not matter to the evolution of $N(t)$. However, this is not the case in the Non-preemptive Greedy policy. For simplicity, in the Non-preemptive Greedy policy, we will view each mixed-type as queue and assume that the tasks of a given mixed-type are served in the FCFS discipline. In other words, if at time $t$ for a given server $s$ we have $z(s,t) = z$, then it serves the task which became of mixed type $z$ the earliest. Note that, in our general model, upon leaving a queue $z$, a task may re-enter the queue at a later point in time. In such a case we consider the arrival time into the queue to be the one corresponding to the latest entry. 

\begin{definition}[Non-preemptive Greedy Policy] In the Non-preemptive Greedy policy, upon completion of an attempt at a task each expert $s$ serving it is assigned an outstanding task such that its success probability $ 1- \psi_s(\tilde z)$ is non-zero. If multiple such tasks exists for a server then it is assigned one which maximizes its success probability. In other words, if an attempt on a task with mixed-type $z$ is completed at time $t$, then for each $s$ such that $z(s,t^-) = z$ we set
$$
z(s,t) \in A'_s(N(t)) \triangleq \arg\min_{\tilde z:N_{\tilde z}(t) > 0, \psi_s(\tilde z)<1} \psi_s(\tilde z),
$$
where ties are broken uniformly at random. 
If no such task exists, \ie, if $A'_s(N(t))$ is empty, then the server stays idle till such a task arrives and starts serving it upon arrival. Further, the tasks with a given mixed-type are served in the FCFS discipline as described above. 
\end{definition}

 For the Non-preemptive Greedy policy, owing to the complexity of the underlying Markov chain, we provide below a rather weak condition for instability which is nonetheless sufficient to establish its sub-optimality. See the Appendix for its proof. 

\begin{proposition}\label{prop:NonPreemptiveGreedy}
Suppose that the Asymmetric($a$) system is stabilizable, \ie, $\lambda < \min\left\{3a/(a+1), 2a \right\}$. Then, under the Non-preemptive Greedy policy, the Asymmetric($a$) system is unstable if we have $\lambda^2(8a^{-1} + 1) +  \lambda (8a^{-1}  - 14) -16 > 0$.  
\end{proposition}

In particular, the above proposition implies that for $a = 1/2$ the throughput of the Asymmetric($a$) system under the Non-preemptive Greedy policy is less than $0.914$, which is sub-optimal. Recall that the optimal throughput for this value of $a$ is $1$.

\section{Experimental Results}
\label{sec:exp}

\begin{table*}
{\normalsize
\begin{center} \caption{Skills of experts estimated by using data from the Math.Stack-Exchange Q\&A platform. The success probabilities with values larger than $35\%$ are highlighted in bold.}
\begin{tabular}{ |c||c||c||c||c||c||c||c||c||c||c|  } 
 \hline
 \multicolumn{11}{|c|}{Expert Clusters} \\
 \hline
 \textbf{Tags} & $1$ & $2$ & $3$ & $4$ & $5$ & $6$ & $7$ & $8$ & $9$ & $10$ \\
 \hline
calculus & .32 & \textbf{.39} &.30 &\textbf{.35} &\textbf{.37} &\textbf{.47} &.28 &.16 &.26 &\textbf{.41} \\
real-analysis & .17 &\textbf{.41} &.25 &.32 &.23 &\textbf{.49} &\textbf{.40} &.10 &.10 &\textbf{.44} \\
linear-algebra &\textbf{.46} &.29 &.05 &\textbf{.36} &.14 &\textbf{.48} &.26 &.31 &.07 &\textbf{.43} \\
probability & .07 &\textbf{.49} &.02 &.33 &.02 &\textbf{.50} &.06 &.02 &\textbf{.46} &.04 \\
abstract-algebra & .02 &.05 &.03 &.32 &.02 &\textbf{.38} &.23 &\textbf{.50} &.01 &.27 \\
integration &.09 &\textbf{.43} &.05 &.19 &\textbf{.44} &\textbf{.45} &.03 &.01 &.06 &\textbf{.37} \\
sequences-and-series &.05 &.32 &.16 &.31 &.20 &\textbf{.45} &.09 &.04 &.06 &.33 \\
general-topology & .02 &.10 &.03 &.16 &.02 &\textbf{.43} &\textbf{.50} &.07 &.02 &.31 \\
combinatorics &.03 &.14 &.06 &\textbf{.43} &.04 &\textbf{.37} &.02 &.06 &.19 &.05 \\
matrices &.27 &.15 &.02 &.31 &.02 &\textbf{.44} &.06 &.11 &.02 &.34 \\
complex-analysis & .02 &.19 &.08 &.16 &.14 &\textbf{.50} &.09 &.05 &.01 &\textbf{.44} \\
 \hline
 \hline
 \textbf{Size} & 165 & 188 & 313 & 200 & 179 & 183 & 231 & 187 & 178 & 176 \\
 \hline
\end{tabular} \label{tab:stack}
\end{center}
}
\end{table*}

In this section, we present our empirical results obtained by using data from Math.Stack-Exchange Q\&A platform. In this platform, users post tagged questions that are answered by other users. Upon resolution of the question, the asker may reveal which of the submitted answers resolved the question. We will use this data to estimate the success probabilities of experts in answering questions, and use these parameters in simulations to compare the throughputs that can be achieved by greedy, random, and backpressure policies. As we will see, a substantially larger throughput can be achieved by backpressure policy than greedy and random.


\paragraph{Dataset}  The dataset consists of around $702,286$ questions and $994,138$ answers. It was retrieved on February 2nd, 2017. The top $11$ most common tags are given in Table~\ref{tab:stack} in decreasing order of popularity. Among these tags, the most common is \emph{calculus} which covers $61,184$ questions, and the least common is \emph{complex analysis} which covers $22,813$ questions. In our analysis, we used only questions that are tagged with at least one of the $11$ most popular tags, which amounts to a total of $381,239$ questions and $544,267$ answers.

\paragraph{Estimated skill sets} The success probabilities of answering questions are estimated as follows. For a given user-tag pair, the success probability is estimated by the empirical frequency of the accepted answers by this user for questions of given tag, conditional on that the user had at least $5$ accepted answers for questions of the given tag, and otherwise we estimate the success probability is set to be equal to zero. These success probabilities are estimated for $2,000$ users with the most accepted answers. Among these users, the user with the most accepted answers had $4,665$ accepted answers, and the user with the least number of accepted answers had $13$ accepted answers. There were $712$ users which had more than $50$ answers accepted. In order to form clusters of users with similar success probabilities for different tags, we ran the k-means clustering algorithm. 

The estimated success probabilities are shown in Table~\ref{tab:stack}. The columns correspond to different centroids of the clusters and give average success probabilities for different tags. In the bottom row, we give the sizes of the corresponding clusters. For instance, the $165$ persons in cluster $1$ have on average $32\%$ of their \emph{calculus}, and $46\%$ of their \emph{linear algebra} answers accepted. 

There is a pronounced heterogeneity in user expertise. We highlighted in bold the success probabilities with values larger than $35\%$. A subset of users, namely cluster $6$, have high success probabilities at all topics whereas the users in the other clusters have high success rate at a subsets of topics. 

{\em Estimating $\pi_z$} 
There is a prevalence of questions with different combinations of tags, that is, mixed types. When a question arrives with multiple tags, we associated with it a mixed-type which is the uniform distribution across the associated tags. We kept only those combinations of tags that occur for at least $1\%$ of the total number of questions. This results in $16$ tag combinations among which $11$ are singletons and $5$ are a combinations of $2$ tags. These are the mixed types $z$ with positive $\pi_z$, we set $\pi_z = 0$ for all other mixed types. From among the questions with these $16$ mixed-types, the fraction of questions which belong the mixed-type $z$ is the estimated for $\pi_z$.  
We observed that roughly $19\%$ of the questions are tagged with multiple tags, showing the relevance of our model. 


\paragraph{Simulation setup}
We assumed that the experts have unit service rates. We make this approximation as we do not have the information about times at which experts begin to respond a question. We examined the system for increasing values of task arrival rates. We simulate our CTMC via a custom discrete event simulator. 

We implement the Backpressure($\mathcal{Y}$) policy where the set $\mathcal{Y}$ consists of all $11$ pure types, the $5$ most frequently seen mixed types upon arrival as described above, and the mixed types which result from an attempt by an expert exactly once. Note that a task belonging to a pure type can be attempted upon multiple times without changing its type.  We thus have $|\mathcal{Y}| = 16 + 5 \cdot 10 = 66$. Our choice of $\mathcal Y$ is a result of a compromise between performance and complexity. Choosing a larger set of $\mathcal Y$ may increase the stability region by a small fraction, but may significantly increase the complexity of the Backpressure($\mathcal{Y}$) policy. 

Further, while serving the tasks in $X$, instead of choosing tasks at random, we choose tasks greedily, \ie, each server is assigned a task in $X$ which maximizes its probability of success. Empirically, this improves the performance over random selection of tasks in $X$.

In the following, we will use the short hand `greedy policy' for the Preemptive Greedy policy, and `backpressure policy' for the Backpressure($\mathcal{Y}$) policy. 





 
\paragraph{Performance comparison of different policies} 

\begin{figure*}
\hspace{-5mm}  \includegraphics[width=0.55\linewidth]{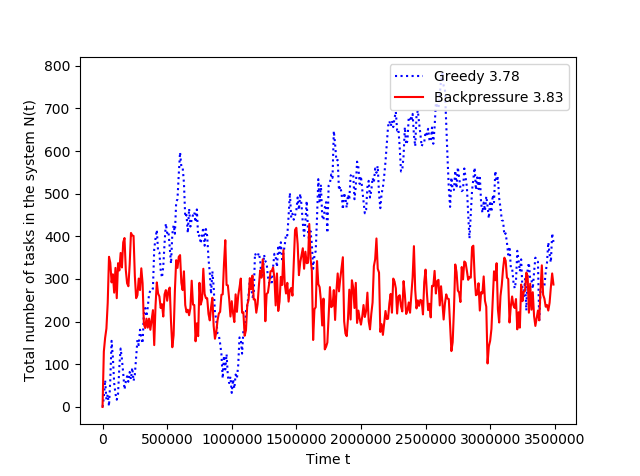} \hspace{-5mm}
 \includegraphics[width=0.55\linewidth]{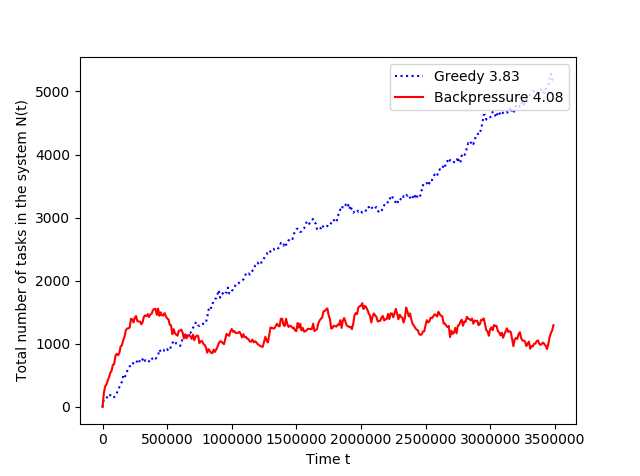}
  \caption{Total number of tasks in the system over time for the greedy and backpressure policy. The task arrival rates are as indicated in the figures.}
  \label{fig::Simulation}
\end{figure*}

In the following, we will use the short hand `greedy policy' for the Preemptive Greedy policy, and `backpressure policy' for the Backpressure($\mathcal{Y}$) policy. In Figure \ref{fig::Simulation} we plot the time-evolution of the total number of active tasks in the system for the greedy policy and the backpressure policy at the respective arrival rates $3.78$ and $3.83$ (Figure \ref{fig::Simulation} left), and also at the arrival rates $3.83$ and $4.08$ (Figure \ref{fig::Simulation} right). In Figure \ref{fig::Simulation} left, both the policies are stable. Yet, the sample path under the backpressure policy is more steady than that under greedy policy, which is an added advantage to its throughput optimality.  In Figure \ref{fig::Simulation} right, while the greedy policy is unstable at $\lambda =3.83$, the backpressure policy is stable even at $\lambda = 4.08$ and thus significantly outperforms the greedy policy. 

\begin{figure}\centering
  \includegraphics[width=0.7\linewidth]{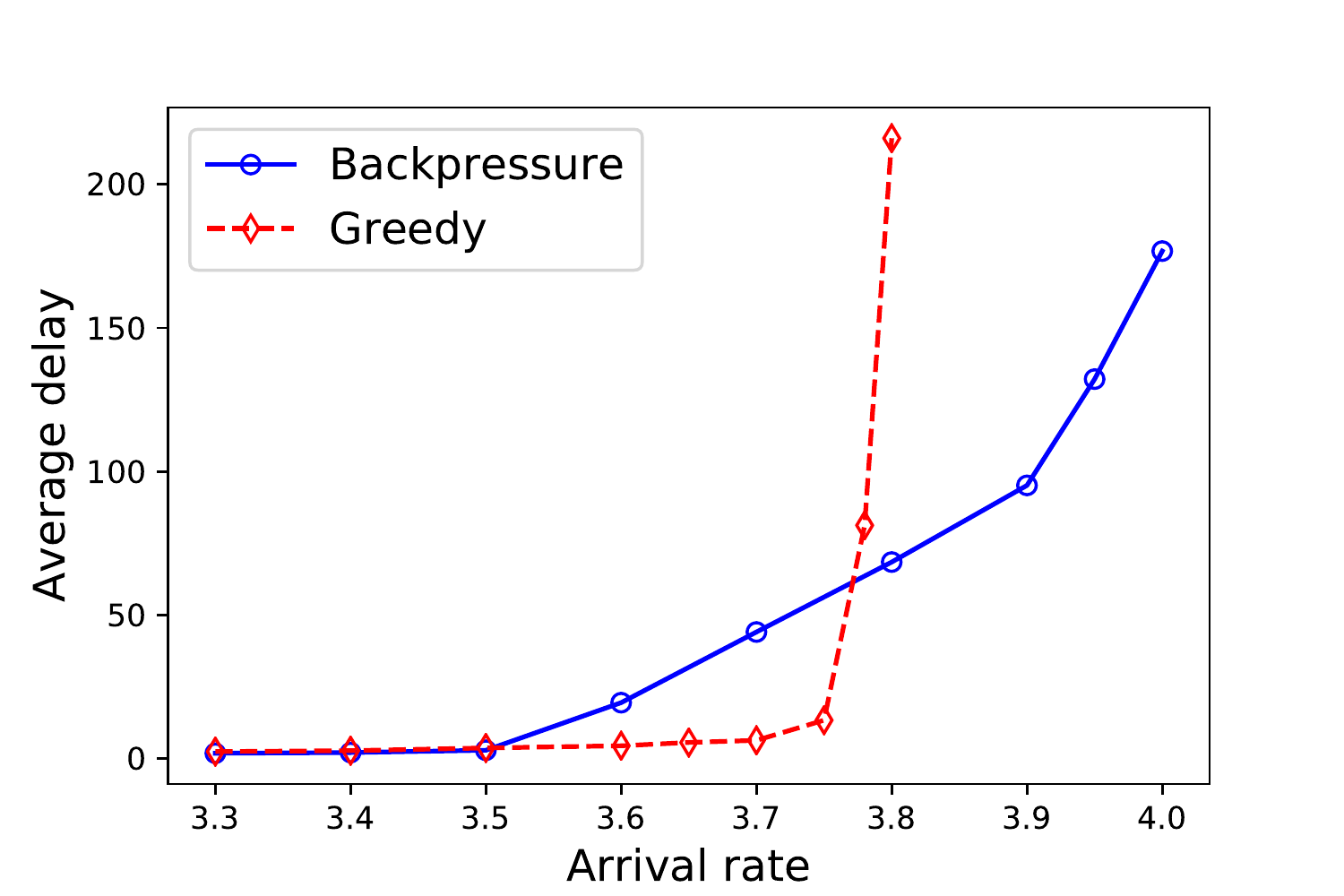}
  \caption{Comparison of performance under greedy and backpressure policies.}
  \label{fig::Arrival}
\end{figure}

In Figure \ref{fig::Arrival} we plot the average delay (sojourn time) of tasks in the system against the task arrival rates. The average delay is computed by first computing the time-averaged number of tasks in the system and then applying Little's law. We observe that the task arrival rates at which random (not shown in the plot), greedy, and backpressure policies become unstable are approximately equal to $2.2$, $3.82$, and $4.10$, respectively. Thus, the backpressure policy achieves throughput improvement of about $8\%$ over the greedy policy. 

\revised{The backpressure policies marginally outperforms greedy in terms of average delay at the low loads, and significantly at high loads. However, observe that at the moderate loads the greedy policy outperforms the backpressure policy. The reason for this is as follows. The backpressure policy achieves throughput optimality by building gradients (in the form of weights) at the large loads which guide system operation. At moderate loads the queue lengths are small and the associated gradients are not very meaningful. This is similar in principle to the well known poor performance of backpressure policy at lower loads in multihop wireless networks, see \cite{YSR11}. Designing policies which perform well at all loads is an interesting avenue for future research. 
}


\revised{
\section{General Feedback Structure}
\label{sec:GeneralFeedback}

The model described in Section~\ref{sec:system} allows for only binary feedback, in the form of success and failure. Upon success a task leaves the system, whereas upon failure, the fact of failure is used to reduce uncertainty in the true-type of the task. In this section we generalize the feedback structure as follows. Upon success a task leaves the system, as in the earlier model. However, upon failure, a server may additionally provide a feedback $f$ from a countable set of possible feedbacks $F$. Let the $\beta_{s,c}(f)$ be the probability that for a task of true type $c\in C$, server $s$ provides a feedback $f$ upon failure. Thus, for each $s$ and $c$, $\beta_{s,c} = (\beta_{s,c}(f): f\in F)$ is a probability mass function. We assume that $\beta_{s,c}$ for each $s$ and $c$ is known. In practice, it needs to be learned. 

In this setting, if an attempt by a server $s$ on a task of mixed type $z$ results into a failure and if the feedback provided by the server is $f$ then the task's new mixed-type, denoted by $\phi_s(z,f)$, is the resulting posterior distribution, namely,
$$\phi_s(z,f) = \left(\frac{z_c(1-p_{s,c})\beta_{s,c}(f)}{\xi_s(z,f)}\right)_{c \in C},$$
where $\xi_s(z,f)$ is the probability that the task for mixed type $z$ results into failure upon an attempt by server $s$ and receives feedback $f$, \ie, $$\xi_s(z,f) = \psi_s(z) \sum_{c} z_c \beta_{s,c}(f).$$
We again assume that, for each $s$ and $f$, $\mathcal Z$ is closed under $\phi_s(\cdot,f)$.

Along the lines of the development of stability conditions in Section~\ref{sec:stability}, we obtain below the necessary and sufficient conditions for stability. Again, we let $\nu_{s,z}$ represent the flow of tasks of mixed-type $z$ served by expert $s$. In developing the new flow conservation constraints we now account for the more general feedback structure. The capacity constraints remain identical. 

\begin{theorem}\label{thm:OptimalStabilityGeneral}
Suppose there exists $s$ such that  $\min_c p_{s,c} >0$. If there exist non-negative real numbers $\nu_{s,z}$ for each $s\in S$ and each $z \in \mathcal Z$, and positive real numbers $\delta_s$ for each $s\in S$ such that the following hold:
\begin{align}
\forall z \in \mathcal Z, \quad \lambda \pi_z + \sum_{\substack{s \in S, f\in F, \\ z' \in \phi^{-1}_s(z,f)}} \nu_{s,z'} \xi_s(z',f) = & \sum_{s\in S} \nu_{s,z}, \\
\forall s \in S, \quad \sum_{z \in \mathcal Z} \nu_{s,z}  + \delta_s \le  \mu_s, 
\end{align}
then there exists a policy under which the system is stable. If there does not exist non-negative real numbers $\nu_{s,z}$ for $s\in S$, $z \in \mathcal Z$ and non-negative real numbers $\delta_s$ for $s\in S$ such that the above constraints hold, then the system cannot be stable.  
\end{theorem}

A stabilizing policy is again obtained by finding a finite set $\mathcal Y$ such that the overall arrival rate into $\mathcal Z \backslash \mathcal Y$ is small, and using a backpressure policy policy for congestion control. More formally, recall the definitions of $(\tilde X_z)_{z\in \mathcal Z}$, $X$, and $(\tilde N_z)_{z\in \mathcal Y}$ from Section~\ref{sec:stability}. Consider the following policy. 

\begin{definition}[Modified Backpressure($\mathcal Y$) policy]\label{def:MBPy} 
For each $s\in S, z\in \mathcal Y$ let 
$$w_{s,z}(\tilde N,X) = 
\tilde N_z  - \sum_{f:\phi_s(z,f) \in\mathcal  Y } \xi_s(z,f) \tilde N_{\phi_s(z,f)} \ - \  X \!\! \sum_{f:\phi_s(z,f) \notin\mathcal  Y } \xi_s(z,f).$$
For a given $(\tilde N,X)$, let 
$$ B_s(\tilde N,X) = \arg\max_{z'\in \mathcal Y: \tilde N_{z'} > 0} w_{s,z}(\tilde N,X).$$
If $$\sum_s \mu_s \max_{z\in \mathcal Y: \tilde N_z > 0} w_{s,z}(\tilde N,X) \ge X  \min_{c\in C} \sum_s \mu_s  p_{s,c}  $$ then each expert chooses a task in $\tilde N_z$ where $z\in B_s(\tilde N,X) \subset \mathcal Y$ with ties broken arbitrarily.
Else, each expert serves a task in $X$ chosen uniformly at random. 
\end{definition}

Again, using the Lyapunov function $L(\tilde N, \tilde X) =   \sum_{z\in \mathcal Y} \tilde N_z^2 + X^2,$ and the arguments identical to the proof of Theorems~\ref{thm:OptimalStability} and \ref{thm:BP} in the Appendix but with appropriate changes, it follows that there exists a finite subset $\mathcal Y$ of $\mathcal Z$ such that the policy Backpressure($\mathcal Y$) stabilizes the system if the necessary conditions are satisfied. Further, the converse statement of the theorem follows from system ergodicity. We omit details for brevity.

}


\section{Related Work}
\label{sec:related}

Bayesian Active Learning (see \cite{GKR10,JCK14,ChK13,FuK16}) aims at learning true hypothesis by adaptively selecting sequence of experiments. In \cite{ChK13} labels are obtained for a batch of items at a time. In \cite{FuK16} a stream based budgeted setting is considered where a finite number of items arrive in a random order. In contrast we allow infinite stream of tasks and are interested in maximizing the task resolution throughput under capacity constraints at the servers. The crowdsourcing works such as \cite{KOS14,Shah15,ZCZ16,Gao16}  consider task assignment problems for classification with unknown ground truths, however they consider a static model.  In \cite{MaX16} the labeling tasks arrive dynamically and their exit is tied to the expert allocation decisions, in that a task leaves once the probability of error in the label estimate falls below a threshold.

Our work is also broadly related to that of \emph{multi-arm bandits}, \eg, see \cite{LaR85,ACF02,GGW11,Buc12,AgG12} and citations therein, in the sense of optimizing the trade-off between exploration, to learn job types, and exploitation, to optimize task performance. It also has some relation with \emph{collaborative filtering systems} such as those studied in \cite{KlS03,Kls04,SHG09}, which can be interpreted as expert-task systems where success probabilities admit a low-rank matrix structure. Unlike our work, there good matches are inferred from observed assignments of tasks to experts, which are according to a given statistical model, and  there are no resources constraints imposed on the experts.

A related line of work is that on \emph{stochastic online matching}, \eg, see  \cite{MeP12,MWZ15,Ho12}. The stochastic online matching can be interpreted as a task-expert system where each expert is associated with a budget constraint that allows to solve at most one task. 
Unlike our work where the task types are uncertain, uncertainty in these models come from the arbitrariness of the future task arrivals and the monotonically decreasing available resource budgets. 

Another related literature is that of \emph{constrained queueing systems}, where arriving tasks are to be served by heterogeneous servers subject to resource constraints, \eg, see \cite{TaE92,NMR03,MaS04,GNT06,YSR11,BSS11,ASK12,JJS13,CYL15,MBS16,SVK16}. The goal is to efficiently utilize server resources while providing good performance in servicing tasks, \eg, optimizing task delays.
Our matching policy is of a flavor similar to the stability-optimal backpressure policy first proposed in \cite{TaE92}. A setting close  to ours is the one studied in \cite{SVK16} for \emph{routing queries in peer-to-peer networks}. Here, the types of the queries are known but the locations of nodes where the queries may by successfully resolved are uncertain. More technically, we associate queues with each prior distribution which may be infinite in number. This makes the stability analysis much more challenging. Another related work is that on \emph{scheduling flexible servers}, \eg, see \cite{MaS04,MBS16}, which allows for tasks of different types and servers of different skills. It has been established that a so called max-weight policy is optimal in a heavy traffic regime. 
The main difference from our work is that all these works assume that the task types are known. 

In \cite{BiM15}, the authors considered a task-expert system where task types are of two difficulty levels (hard or easy) and expert skills are of two levels (senior or junior). Seniors may serve any task, but juniors may only serve easy tasks. The hardness of each task is unknown upon arrival. 
In comparison, we allow for much more generality with respect to the heterogeneity of skills of experts. In their model, a task upon service can only become progressively harder, which amounts to a feed-forward system, unlike our model.

The work in \cite{JKK17} considers a model where the job types are known but the expert types are unknown. They consider the problem of matching while simultaneously learning the expert types. A key idea is to use a shadow price which simultaneously accounts for resource utilization and type uncertainties. They consider an asymptotic regime where each expert is allowed to work on a large number of tasks, a vanishingly small amount of which could be used to accurately learn the expert types, and the rest can be served optimally. In the limit, the learning aspect is decoupled form the expert utilization, and it is thus different from our work.

\section{Conclusion}
\label{sec:conc}

We studied matching of tasks and experts in a system with uncertain task types. We established a complete characterization of the stability region of the system, i.e. the set of task arrival rates that can be supported by a matching policy such that the expected number of tasks waiting to be served is finite. We showed that any task arrival rate in the stability region can be supported by a back-pressure matching policy. We also compared with two baseline matching polices, and identified instances under which there is a substantial gap between the maximum task arrival rates that can be supported by these policies and that of the optimum back-pressure matching policy.

There are several interesting directions for future research. First, for the case when task types are unknown, it is of interest to consider matching policies that optimize different kinds of performance objectives, such as, for example, minimizing the long-run average of a function of task waiting times. Second, much remains to be said about matching policies for the case when both task types and the skills of experts are unknown. 



\section{Proofs}\label{sec:proofs}

\subsection{Proof of Theorem~\ref{thm:OptimalStability} and Theorem \ref{thm:BP}} \label{proof:OptimalStability}
We first show stability under sufficient conditions provided in the statement of Theorem~\ref{thm:OptimalStability}. In the process, we prove Theorem~\ref{thm:BP}. 
 
In constrained queueing systems, \eg, see \cite{TaE92,GNT06}, a standard approach towards proving stability of a backpressure type policy is to design a `static' policy using flow variables $(\nu_{sz})_{s,z}$ and the slacks $(\delta_s)_s$ which provides a fixed service rate to each queue $N_z$ such that its drift is sufficiently negative for each. However, in our setup the total number of queues $(N_z)_{z\in \mathcal Z}$ could be countable, while the total available slack is finite. Thus, it is not possible to design a static policy such that the drift in each individual queue is bounded from above by a negative constant. This is unlike any finite-server queueing system considered in the previous literature. 

We thus take a different approach, which can be explained roughly as follows. Since the total exogenous arrival rate $\lambda$, and the total endogenous arrival rate, \ie\ arrival into a queue due to failure at another queue, are both finite (they are bounded from above by $\sum_s \mu_s$),  there exists a finite set $\mathcal Y \subset \mathcal Z$ such that the total arrival rate into $\mathcal Z \backslash \mathcal Y$ is less than $\min_{c\in C}  \sum_{s\in S} \frac{\delta_s}{4} p_{s,c}$. Each task which enters a queue $N_z$ where $z \in \mathcal Z \backslash \mathcal Y$ is instead sent to a virtual queue $X$, and stays there until there is a success. If $X$ is `large' compared to the other queues then all the servers focus on $X$. The finite number of remaining queues are operated via a backpressure policy which accounts for the `expected backlog' seen in these queues.  

More formally, consider $(\nu_{s,z})_{s,z}$ and positive constants $(\delta_s)_s$ as postulated in the theorem. Without loss of generality, assume that there exists a constant $0<\epsilon<1$ such that $\delta_s = \epsilon \mu_s$ for each $s\in S$. Let $\mathcal Y$ be a finite subset of $\mathcal Z$ such that 
\begin{equation} \label{eq:ChoiceOfY}
\sum_{z \in \mathcal Z \backslash \mathcal Y} \left( \lambda \pi_z +  \sum_{s \in S} \sum_{z' \in \phi^{-1}_s(z) \cap \mathcal Y} \nu_{s,z'} \psi_s(z') \right)
\le \min_{c\in C}  \sum_{s\in S} \frac{\delta_s}{4} p_{s,c}.
\end{equation}
Since $\lambda + \sum_{s\in S,z\in \mathcal Z} \nu_{s,z} \le 2 \sum_s \mu_s$, such a $\mathcal Y$ exists.

Let $X$ be the number of tasks in the system which are or have been in past of type $z \in \mathcal Z \backslash \mathcal Y$. Once a task enters queue $X$ it does not leave it until success. There may be tasks in it with mixed-type in $\mathcal Y$. Note, our policy will depend on $X$ and thus $(z(s,t))_s$ will not be $N(t)$ measurable. In turn, $N(t)$ will not be a CTMC. For $z \in \mathcal Y$, let $\tilde X_z$ and $\tilde N_z$ be the tasks of mixed-type $z$ which have and have not had mixed-type in $ \mathcal Z \backslash \mathcal Y$. Also, for convenience  for each $z \in \mathcal Z \backslash \mathcal Y$, let $\tilde X_z$ be the tasks of mixed-type $z$, \ie, $N_z = \tilde X_z$ for each $z \in \mathcal Z \backslash \mathcal Y$. We now formally define $\sigma\left((\tilde X_z)_{z\in \mathcal Z},(\tilde N_z)_{z\in \mathcal Y}\right)$-measurable backpressure policy. Thus, $\left((\tilde N_z)_{z\in \mathcal Y},(\tilde X_z)_{z\in \mathcal Z}\right)$ is a CTMC. 

We now show stability of the system under this policy for Backpressure($\mathcal Y$) as given in Definition~\ref{def:BPy}. 
Below we will assume that the ties in selecting $z$ from $B_s(\tilde N,X)$ are broken uniformly at random for simplicity of exposition. The proof can be easily extend to any other tie breaking approach. 
Consider the following Lyapunov function. 

$$
L(\tilde N, \tilde X) = \sum_{z\in \mathcal Y} \tilde N_z^2 +  \left(\sum_{z\in \mathcal Z} X_z\right)^2 =  \sum_{z\in \mathcal Y} \tilde N_z^2 + X^2.
$$ 

 For each $t$, let $t+\tau(t)$ be the time at which the first event (arrival or completion of a response) occurs after time $t$. Clearly, $\tau(t)$ is a stopping time.
Further, let $\tau_{\tilde n, \tilde x}(t) = E[\tau(t)| (\tilde N(t), \tilde X(t)) = (\tilde n, \tilde x)]$.  

Let
$$
D(\tilde n, \tilde x) \define \frac{1}{\tau_{\tilde n, \tilde x}} E\left[L(\tilde N(t+\tau), \tilde X(t+\tau))  - L(\tilde N(t), \tilde X(t))\big| \tilde N(t) = \tilde n, \tilde X(t) = \tilde x \right].$$
$D(\tilde n, \tilde x)$ is called drift in state $n$. We would like to show that there exists a positive integer $K$ and positive constant $\epsilon$ such that 

$$D(\tilde n, \tilde x) \le -\epsilon  \;\; \forall (\tilde n, \tilde x) \st  \max(|\tilde n|_\infty, x) \ge K.$$


Let for each $s\in S$ and $z \in \mathcal Y$ let 
$$\nu^*_{s,z} =\indic{x  \min_{c\in C} \sum_s \mu_s  p_{s,c}  > \sum_s \mu_s \max_{z\in \mathcal Y: \tilde n_z > 0} w_{s,z}(\tilde n,x)}  \indic{z\in  B_s(n)} \frac{1}{|B_s(n)|}.$$  Then, one can check that  
\begin{multline*}
$$\frac{1}{\tau_{\tilde n, \tilde x}} E[\tilde N_z(t+\tau)^2  - \tilde N_z(t)^2 \big| \tilde N(t)  = \tilde n, \tilde X(t) = \tilde x ]  \\ =   (2\tilde n_z +1) \left(\lambda \pi_z +  \sum_{s \in S} \sum_{z' \in \phi^{-1}_s(z) \cap \mathcal Y} \nu^*_{sz'} \psi_s(z') \right)   +  (-2\tilde n_z +1)  \sum_s \nu^*_{s,z}.$$
\end{multline*}

Further, let 
$$\nu^* = \indic{x  \min_{c\in C} \sum_s \mu_s  p_{s,c}  > \sum_s \mu_s \max_{z\in \mathcal Y: \tilde n_z > 0} w_{s,z}(\tilde n,x)}  .$$ Then, we have that 
\begin{multline*}
\frac{1}{\tau_{\tilde n, \tilde x}}  E[X(t+\tau)^2  - X(t)^2 \big| \tilde N(t)  = \tilde n, \tilde X(t) = \tilde x ] \\  \le  (2x+1) \sum_{z \in \mathcal Z \backslash \mathcal Y} \left( \lambda \pi_z +  \sum_{s \in S} \sum_{z' \in \phi^{-1}_s(z) \cap \mathcal Y} \nu^*_{s,z'} \psi_s(z') \right)  + (-2x+1)  \nu^* \min_c\sum_s \mu_s p_{s,c}. 
\end{multline*}

Thus, we get 
\begin{multline*}
D(\tilde n, \tilde x) \le \sum_{z\in \mathcal Y} (2\tilde n_z +1) \left(\lambda \pi_z +  \sum_{s \in S} \sum_{z' \in \phi^{-1}_s(z) \cap \mathcal Y} \nu^*_{s,z'} \psi_s(z') \right)   +  (-2\tilde n_z +1)  \sum_s \mu_s \nu^*_{s,z} \\
+ (2x+1) \sum_{z \in \mathcal Z \backslash \mathcal Y} \left( \lambda \pi_z +  \sum_{s \in S} \sum_{z' \in \phi^{-1}_s(z) \cap \mathcal Y} \nu^*_{s,z'} \psi_s(z') \right)  + (-2x+1) \nu^* \min _c \sum_s \mu_s p_{s,c}  .
\end{multline*}

Upon arranging terms, we obtain 
\begin{multline*}
D(\tilde n, \tilde x) \le \sum_{z\in \mathcal Y} 2\tilde n_z \left(\lambda \pi_z +  \sum_{s \in S} \sum_{z' \in \phi^{-1}_s(z) \cap \mathcal Y} \nu^*_{sz'} \psi_s(z') -\sum_s \nu^*_{s,z} \right)  \\
+ 2x \left( \sum_{z \in \mathcal Z \backslash \mathcal Y} \left( \lambda \pi_z +  \sum_{s \in S} \sum_{z' \in \phi^{-1}_s(z) \cap \mathcal Y} \nu^*_{sz'} \psi_s(z') \right) -    \nu^* \min _c \sum_s \mu_s p_{s,c} \right) \\
+  \left (\lambda +  \sum_{z\mathcal Z}  \sum_{s \in S} \sum_{z' \in \phi^{-1}_s(z) \cap \mathcal Y} \nu^*_{s,z'} \psi_s(z') + \sum_{z\in \mathcal Y} \sum_s  \nu^*_{s,z}  +    \nu^* \min _c \sum_s \mu_s p_{s,c}  \right).
\end{multline*}
The last of the above three terms can be bounded by a constant, say $\alpha_1 = 10 \sum_s \mu_s$. 
For each $s\in S$ and $z \in \mathcal Y$ let $\hat\nu^*_{s,z} = (\mu_s-3\delta_s/4) \nu^*_{s,z}$ and $\tilde\nu^*_{sz} = (\delta_s/4)  \nu^*_{s,z}$. Further, let $\hat \nu^* =  \min _c \sum_s  (\mu_s - 3\delta_s/4)p_{s,c}  \nu^*$ and $\tilde \nu^* = \min _c \sum_s (\delta_s/4) p_{sc}  \nu^*$.
Then,
\begin{multline} \label{eq:D1}
D(\tilde n, \tilde x) \le \sum_{z\in \mathcal Y} 2\tilde n_z \left(\lambda \pi_z +  \sum_{s \in S} \sum_{z' \in \phi^{-1}_s(z) \cap \mathcal Y} \hat\nu^*_{sz'} \psi_s(z') -\sum_s \hat\nu^*_{s,z} \right)  \\
+ 2x \left( \sum_{z \in \mathcal Z \backslash \mathcal Y} \left( \lambda \pi_z +  \sum_{s \in S} \sum_{z' \in \phi^{-1}_s(z) \cap \mathcal Y} \hat\nu^*_{s,z'} \psi_s(z') \right) - \hat \nu^*\right) + \alpha_1 \\
+ \sum_{z\in \mathcal Y} 2\tilde n_z \left( \sum_{s \in S} \sum_{z' \in \phi^{-1}_s(z) \cap \mathcal Y} \tilde\nu^*_{s,z}\psi_s(z') -\sum_s\tilde\nu^*_{sz} \right) + 2x \left( \sum_{s \in S} \sum_{z' \in \phi^{-1}_s(z) \cap \mathcal Y} \tilde \nu^*_{s,z'} \psi_s(z') - \tilde \nu^* \right).
\end{multline}

Consider the following lemma. Its proof is given in Section~\ref{proof:OptimalWeights}.

\begin{lemma}\label{lemma:OptimalWeights}
Recall the $(\nu_{s,z})_{s,z}$ as postulated by the theorem. For $\Theta = (\theta_{s,z})_{s\in S, z\in \mathcal Y} \cup (\theta) $, where $\theta$ and $\theta_{s,z}$ for each $s,z$ are reals, let
\begin{multline*}
f(\Theta) =  \sum_{z\in \mathcal Y} 2\tilde n_z \left(\lambda \pi_z +  \sum_{s \in S} \sum_{z' \in \phi^{-1}_s(z) \cap \mathcal Y} \theta_{sz'} \psi_s(z') -\sum_s \theta_{s,z} \right)  
 \\ + 2x \left( \sum_{z \in \mathcal Z \backslash \mathcal Y} \left( \lambda \pi_z +  \sum_{s \in S} \sum_{z' \in \phi^{-1}_s(z) \cap \mathcal Y} \theta_{sz'} \psi_s(z') \right) - \theta \right). 
\end{multline*}
 Then, 
 $$f\left((\hat\nu^*_{s,z})_{s\in S, z\in \mathcal Y} \cup \hat \nu^* \right) \le f\left((\nu_{s,z})_{s\in S, z\in \mathcal Y} \cup \{\min_c \sum_{s\in S} (\delta_s/4) p_{sc}\}\right).$$
\end{lemma}


%
From definition of $ \nu_{s,z} $, we get that the first term in $f(\Theta)$ for $\Theta = (\nu_{s,z})_{s\in S, z\in \mathcal Y} \cup (\min_c \sum_{s\in S} (\delta_s/4) p_{sc})$ is equal to $0$, and, from \eqn{eq:ChoiceOfY} we have that the second term in it is less than or equal to 0. 

Thus, we have that $f\left((\hat\nu^*_{s,z})_{s\in S, z\in \mathcal Y} \cup \hat \nu^* \right) \le 0$. 
From \eqn{eq:D1} we in turn obtain
$$D(\tilde n, \tilde x) \le 
\alpha_1 
+ \sum_{z\in \mathcal Y} 2\tilde n_z \left( \sum_{s \in S} \sum_{z' \in \phi^{-1}_s(z) \cap \mathcal Y} \tilde\nu^*_{s,z}\psi_s(z') -\sum_s\tilde\nu^*_{s,z} \right) \\ + 2x \left(\sum_{s \in S} \sum_{z' \in \phi^{-1}_s(z) \cap \mathcal Y} \tilde \nu^*_{s,z'} \psi_s(z') -  \tilde \nu^* \right) . $$

%
%
%
%

Fix $\epsilon > 0$. We now show that there exists a positive integer $K$ such that if $x> K$ or if $|\tilde n|_\infty > K$ then $D(\tilde n, \tilde x) \le - \epsilon$. Upon rearranging terms, we obtain 

\begin{align*}
D(\tilde n, \tilde x) & \le 
 \alpha_1 
- 2\sum_{s \in S} \sum_{z\in \mathcal Y: \phi_s(z) \in \mathcal Y} \tilde \nu^*_{s,z} (\tilde n_z  - \psi_s(z) \tilde n_{\phi_s(z)} ) - 2\sum_{s \in S} \sum_{z\in \mathcal Y: \phi_s(z) \in \mathcal Z\backslash \mathcal Y} \tilde \nu^*_{s,z} (\tilde n_z  - \psi_s(z) X ) - 2x \tilde \nu^* \\
& = \alpha_1 
- 2\sum_{s \in S} \sum_{z\in \mathcal Y} \tilde \nu^*_{s,z} w_{s,z}(\tilde n,x) - 2 \tilde \nu^* x, \\
&= \alpha_1 - \max\left(2\sum_{s \in S} \sum_{z\in \mathcal Y} \tilde \nu^*_{s,z} w_{s,z}(\tilde n,x) , 2 \tilde \nu^* x\right). 
\end{align*}

From the definition of the algorithm we get that
$$D(\tilde n, \tilde x) \le \alpha_1 - x \min_{c\in C}  \sum_{s\in S} \frac{\delta_s}{4} p_{s,c}.$$
Hence, for any $(\tilde n,x)$ such that $x> ( \alpha_1 + \epsilon) \min_{c\in C}  \sum_{s\in S} \frac{\delta_s}{4} p_{s,c}  $, we have $D(\tilde n, \tilde x) \le - \epsilon$. 

We also have that 
$$D(\tilde n, \tilde x) \le \alpha_1 
- 2\sum_{s \in S} \frac{\delta_s}{4}\max_{z \in \mathcal Y} w_{s,z}(\tilde n,x).$$
Thus, 
$$D(\tilde n, \tilde x) \le \alpha_1 
- 2 \left(\min_{s\in S} \frac{\delta_s}{4} \right) \sum_{s \in S} \max_{z \in \mathcal Y} w_{s,z}(\tilde n,x) \le \alpha_1 
- 2 \left(\min_{s\in S} \frac{\delta_s}{4} \right) \max_{z \in \mathcal Y}  \sum_{s \in S} w_{s,z}(\tilde n,x). $$

Now suppose that $x\le \alpha_2 \define ( \alpha_1 + \epsilon) \min_{c\in C}  \sum_{s\in S} \frac{\delta_s}{4} p_{s,c}$. Then, if we are able to show that $\max_{z \in \mathcal Y}  \sum_{s \in S} w_{s,z}(\tilde n,x) \to \infty$ as $|\tilde n|_\infty \to \infty$, then we would have that $D(\tilde n, \tilde x) \le - \epsilon$ a positive integer $K'$ such that $|\tilde n|_\infty > K'$. We now show that, under $x \le \alpha_2$, we have $ \sum_{s \in S}\max_{z \in \mathcal Y} w_{s,z}(\tilde n,x) \to \infty$ as $|\tilde n|_\infty \to \infty$.

Let $z^* \in \arg \max_{z \in \mathcal Y} \tilde n_z$. Then we have
\begin{align*}
 \sum_{s \in S}w_{s,z*}(\tilde n,x) &  \le \sum_s( \tilde n_{z^*} - \psi_s(z^*) \max(\alpha_2, \tilde n_{z^*}) \\
 & = |S| \tilde n_{z^*} - \max(\alpha_2, \tilde n_{z^*}) \sum_s \psi_s(z). 
 \end{align*}
which tends to infinity because
\begin{align*}
\sum_s \psi_s(z^*) &= \sum_s \sum_c z^*_c (1-p_{s,c}) = |S| - \sum_s \sum_c z^*_c p_{s,c} \\
& \le |S| - \max_c \sum_s  z^*_c p_{s,c} \le |S| - \max_c z^*_c \min_c \sum_s p_{s,c} \\
& \le  |S| -  \frac{1}{|C|}  \min_c \sum_s p_{s,c}  < |S|.
\end{align*}

Thus, there exist positive constants $K$ and $\epsilon$ such that if $x> K$ or if $|\tilde n|_\infty > K$ then $D(\tilde n, \tilde x) \le - \epsilon$.

Let $\mathcal A \define \{(\tilde n, \tilde x): \max( |\tilde n|_\infty, x ) \le K\}$. Then, using a variant of Lyapunov-Foster theorem, namely Theorem 8.13 in \cite{Rob03}, we obtain that  
from any state $(\tilde n, \tilde x)$ such that $|\tilde n| + x < \infty$, the expected time to return to $\mathcal A$, \ie, $\tau_{\mathcal A}(\tilde n , \tilde x)$ is finite. Further,
$$T \define \sup_{(\tilde n , \tilde x) \in \mathcal A}  \tau_{\mathcal A}(\tilde n , \tilde x) < \infty.$$

Thus, starting with any state in $\mathcal A$, we return to $\mathcal A$ in a finite expected time. We will be done if we show that expected time to return to state $(0,0)$ is also finite. We do this as follows. Fix a constant $\beta >0$. Since there exists $s$ such that  $\min_c p_{s,c} >0$, we have that for any interval of time of size $\beta$ the probability that no arrival happens in this interval and that a task leaves the system is finite. 

Suppose that system is in a state $(\tilde n , \tilde x) \in \mathcal A$ at time $t=0$. Now consider renewal times $T_i, i=0,1, 2, \ldots$, where $T_0 = 0$ and for each  $i>0$, $T_i$ is defined as follows: $T_i$ is equal to $T_{i-1} + \beta$ if indeed no arrival happens and a task leaves the system in the interval $[T_{i-1}, T_{i-1} + \beta)$, else $T_i$ is the first time of return to $\mathcal A$ after $T_{i-1}$. Clearly $E[T_i] $ since $T$ as defined above is finite. Further probability that a task leaves system in time $T_i - T_{i-1}$ is finite, say $\alpha$. Thus, time for system emptying after first reaching $\mathcal A$ can be upper-bounded by sum of $K$ geometric random variables with rate $\alpha$. Thus expected time to return to state $(0,0)$ is finite. Hence, the system is stable. 

Now suppose that the system is stable. Then, the necessary conditions can be shown to hold by the ergodicity of the system, and letting $\nu_{s,z}$ for each $s,z$ to be the long-term fraction of times a server $s$ attempts a task in $N_z$. 
\qed

\subsection{Proof of Lemma \ref{lemma:OptimalWeights}}\label{proof:OptimalWeights}

Upon rearrangement of terms in the expression of $f(\Theta)$ we obtain
$$f(\Theta)/2 =  - \sum_s \sum_{z\in \mathcal Y: \phi_s(z)\in \mathcal Y}   \theta_{s,z}  (n_z - \psi_s(z') n_{ \phi_s(z)}) -  \sum_s \sum_{z\in \mathcal Y: \phi_s(z)\in \mathcal Z\backslash Y}   \theta_{s,z}  (n_z - \psi_s(z') x) - x \theta. 
$$

By using the definition of weights $w_{s,z}$, we obtain

$$f(\Theta)/2 =  - \sum_s \sum_{z\in \mathcal Y}   \theta_{s,z}  w_{s,z}(\tilde n,x) - x \theta  \ge - \sum_s \left(\max_{z\in \mathcal Y} w_{s,z}(\tilde n,x)  \right) \sum_{z\in \mathcal Y}   \theta_{s,z} - x \theta .$$

 Thus, 

\begin{align*}
&f((\nu_{sz})_{s\in S, z\in \mathcal Y} \cup \{\min_c \sum_{s\in S} (\delta_s/4) p_{sc}\})/2 \\
& \ge - \sum_s \left(\max_{z\in \mathcal Y} w_{s,z}(\tilde n,x) \right) \sum_{z\in \mathcal Y}   \nu_{s,z} - x  \min_c \sum_{s\in S} (\delta_s/4) p_{sc} \\
 &\ge - \sum_s  (\mu_s - \delta_s/2) \max_{z\in \mathcal Y} w_{sz}(\tilde n,x)  -  x  \min_c \sum_{s\in S} (\delta_s/4) p_{s,c}  \\
& \ge -   \indic{ \sum_s \max_{z\in \mathcal Y} w_{s,z}(\tilde n,x)  (\mu_s - 3\delta_s/4) \ge x ( \min_c \sum_s (\mu_s - 3\delta_s/4) p_{s,c}  ) } \sum_s \max_{z\in \mathcal Y} w_{s,z}(\tilde n,x)  (\mu_s - 3\delta_s/4)  \\ & -   \indic{ \sum_s \max_{z\in \mathcal Y} w_{s,z}(\tilde n,x)  (\mu_s - 3\delta_s/4) < x ( \min_c \sum_s (\mu_s - 3\delta_s/4) p_{s,c}  ) } x \min_c \sum_s (\mu_s - 3\delta_s/4) p_{s,c}    \\
& = f\left((\hat\nu^*_{sz})_{s\in S, z\in \mathcal Y} \cup \hat \nu^* \right)/2.
\end{align*}
Hence, the lemma holds. \qed

\subsection{Proof of Theorem \ref{thm:AltBP}}

\newcommand{\EE}{{\mathbf{E}}}
\newcommand{\PP}{{\mathbf{P}}}
\newcommand{\RR}{{\mathbb{R}}}
\newcommand{\CC}{{\mathbb{C}}}
\newcommand{\VAR}{{\mathrm{Var}}}
\newcommand{\dE}{\mathbb {E}}
\newcommand{\dP}{\mathbb{P}}
\newcommand{\dN}{\mathbb {N}}
\newcommand{\dR}{\mathbb {R}}
\newcommand{\setR}{\mathbb {R}}
\newcommand{\dC}{\mathbb {C}}
\newcommand{\cF}{\mathcal {F}}
\newcommand{\dH}{\mathbb{H}}
\newcommand{\cE}{\mathcal {E}}
\newcommand{\cG}{\mathcal {G}}
\newcommand{\cX}{\mathcal {X}}
\newcommand{\cL}{\mathcal {L}}
\newcommand{\cW}{\mathcal {W}}
\newcommand{\cP}{\mathcal {P}}
\newcommand{\cV}{\mathcal {V}}
\newcommand{\cN}{\mathcal {N}}
\newcommand{\cS}{\mathcal {S}}
\newcommand{\cD}{\mathcal {D}}
\newcommand{\cT}{\mathcal {T}}
\newcommand{\cY}{\mathcal {Y}}
\newcommand{\cZ}{\mathcal {Z}}
\newcommand{\cC}{\mathcal {C}}
\newcommand{\Bin}{ \mathrm{Bin}}
\newcommand{\Poi}{ \mathrm{Poi}}
\newcommand{\DTV}{{\mathrm{d_{TV}}}}

\newcommand{\vol}{V}
\newcommand{\sign}{ \mathrm{sign}}
\newcommand{\diag}{ \mathrm{diag}}
\newcommand{\supp}{ \mathrm{supp}}
\newcommand{\bl}{[\hspace{-1pt}[}
\newcommand{\br}{]\hspace{-1pt}]}

\newcommand{\FLOOR}[1]{{{\lfloor#1\rfloor}}} %
\newcommand{\CEIL}[1]{{{\lceil#1\rceil}}} %
\newcommand{\ABS}[1]{{{\left| #1 \right|}}} 
\newcommand{\BRA}[1]{{{\left\{#1\right\}}}} 
\newcommand{\SBRA}[1]{{{\left[#1\right]}}} 
\newcommand{\DOT}[1]{{{\left<#1\right>}}} 
\newcommand{\ANG}[1]{{{\langle#1\rangle}}} 
\newcommand{\NRM}[1]{{{\left\| #1\right\|}}} 
\newcommand{\NRMS}[1]{{{\| #1\|}}} 
\newcommand{\NRMHS}[1]{\NRM{#1}_\text{HS}} 
\newcommand{\OSC}[1]{{{\p(){\mathrm{osc}}{#1}}}} 
\newcommand{\PAR}[1]{{{\left(#1\right)}}} 
\newcommand{\BPAR}[1]{{{\biggl(#1\biggr)}}} 
\newcommand{\BABS}[1]{{{\biggl|#1\biggr|}}} 

\newcommand{\1}{1\!\!{\sf I}}\newcommand{\IND}{\1}
\newcommand{\veps}{\varepsilon}
\newcommand{\si}{\sigma}
\newcommand{\jbf}{{\bf{j}}}
\newcommand{\ibf}{{\bf{i}}}
\newcommand{\ubf}{{\bf{1}}}

\newcommand{\whp}{{w.h.p.~}}
\newcommand{\wop}{{with overwhelming probability}}
\newcommand{\wvhp}{{w.v.h.p.~}}

\newcommand{\BEAS}{\begin{eqnarray*}}
\newcommand{\EEAS}{\end{eqnarray*}}
\newcommand{\BEA}{\begin{eqnarray}}
\newcommand{\EEA}{\end{eqnarray}}
\newcommand{\BEQ}{\begin{equation}}
\newcommand{\EEQ}{\end{equation}}
\newcommand{\BIT}{\begin{itemize}}
\newcommand{\EIT}{\end{itemize}}
\newcommand{\BNUM}{\begin{enumerate}}
\newcommand{\ENUM}{\end{enumerate}}

Suppose that the sufficient conditions as given in Theorem~\ref{thm:OptimalStability} are satisfied. Then, in the proof of Theorem \ref{thm:OptimalStability} we showed existence of a policy such that the system is ergodic.
In fact, since we have a strict slack $\delta_s >0$ for capacity constraint at each server, using proof of Theorem \ref{thm:OptimalStability} we can design a policy for a system which achieves stability even when the server capacities are modified as $\mu'_s = \mu_s - R$, where $0< R < \min_s \delta_s$. Under such a policy,  for each $s\in S$, $z\in\cZ$, $t\ge 1$, let $\mu_{s,z}(t)$ represent the long-term fraction of times a server $s$ attempts a task in $N_z$ which has been attempted $t-1$ times in the past. Then, the following hold. 
\begin{equation}\label{rates_1}
\begin{array}{ll}
i)&\lambda_z=\sum_s \mu_{s,z}(1),\\
ii)&\sum_s \mu_{s,z^-_s}(t)\psi_s(z^-_s)=\sum_s \mu_{s,z}(t+1),\\
iii)&\sum_{t\ge 1}\sum_{z\in\cZ}\mu_{s,z}(t)\le \mu_s,\\
iv)&\exists s_0: \sum_{t\ge 1}\sum_{z\in\cZ}\mu_{s_0,z}(t)= \mu_{s_0}-R, \; R>0.
\end{array}
\end{equation}

The inequalities in \eqn{rates_1} can be strengthened to achieve positive slack for each server's capacity, but \eqn{rates_1} as mentioned is sufficient for our purposes. 
Using existence of $(\mu_{s,z}(t) : s\in S, z \in \cZ)$ which solves \eqn{rates_1}, we now show that, for Backpressure($\epsilon$) policy, provided $\epsilon>0$ has been chosen small enough, the function $L(n):=\sum_{i} n(A_i)^2$ is a Lyapunov function in the sense that its drift is negative, bounded away from 0 except for states $n$ with $\sum_z n_z\le N$ for some threshold $N$. This will imply the announced result by the same arguments as in the proof of Theorem~\ref{thm:BP}.

Let $n=(n_z)$ be given. For each $A_i$ such that $n(A_i)>0$, we pick arbitrarily one point $z_i$ in $A_i$ such that $n_{z_i}>0$. We then define the projection operator $P(z)$ which maps $z$ to $z_i$ if $z\in A_i$. For $z\in A_i$ such that $n(A_i)=0$ we say that $P(z)$ is undefined. We shall also consider for each $z\in\cZ$ the operator
\begin{equation}\label{P_op}
P^t_s(z):=P(\phi_s(P^{t-1}_s(z))).
\end{equation}
This is defined so long as all the involved projections are defined, i.e. the constructed sequence only visits sets $A_i$ with $n(A_i)>0$. We also let $\phi^t_s$ denote the application resulting from $t$ applications of $\phi_s$. 

We now define for each $s,z,t,z_i$ the following rates:
\begin{equation}
\label{rates_2}
\begin{array}{ll}
\nu_{s,z_i}(1;z)&:=\mu_{s,z}(1)\1_{P(z)=z_i},\\
\nu_{s,z_i}(t;z)&:=\left\{\begin{array}{ll}\mu_{s,\phi^t_s(z)}(t)\1_{P^t_s(z)=z_i}&\hbox{ if $P^t_s(z)$ is defined},\\
0&\hbox{otherwise.}
\end{array}
\right.
\end{array}
\end{equation}
Finally, we define the following rates for all $t\le T$, where $\epsilon'$ and  $\beta$ are constants to be specified shortly:
\begin{equation}
\label{rates_3}
\begin{array}{ll}
r_{s,z_i}(1)&:=\sum_{z\in \cZ}\nu_{s,z_i}(1;z),\\
r_{s,z_i}(t)&:=\sum_{z\in \cZ}\nu_{s,z_i}(t;z)+\epsilon'(1+\beta^t)\1_{s=s_0}\1_{n(A_i)>0}.
\end{array}
\end{equation}
We extend the definition of the rates $r_{s,z_i}(t)$ for $t>T$ by induction as follows. First, for $s\ne s_0$ we let $r_{s,z_i}(t)=0$. For server $s_0$, we let
\begin{equation}
\label{rates_4}
r_{s_0,z_i}(T+1):=\sum_s\sum_{z:P^{T+1}_s(z)=z_i}\nu_{s,z_i}(T+1;z)+\epsilon'(1+\beta^{T+1})
\end{equation}
and for $t>T$:
\begin{equation}
\label{rates_5}
r_{s_0,z_i}(t+1)=\sum_{j: P^1_{s_0}(z_j)=z_i}(1-\alpha)r_{s_0,z_j}(t).
\end{equation}
The functions $\psi_s$ are all Lipschitz-continuous. Under the assumption \eqref{superfluous}, it is easily verified that the functions $\phi_s$ are also Lipschitz-continuous. Let $K$ be such that all these functions are $K$-Lipschitz-continuous. 

It is readily established by induction on $t$ that for all $s$, so long as $P^t_s(z)$ is defined, one has
\begin{equation}\label{lipschitz_iterates}
|P^t_s(z)-\phi^t_s(z)|\le \epsilon\frac{K^t-1}{K-1}.
\end{equation}
Indeed, one has
$$
\begin{array}{ll}
|P^t_s(z)-\phi^t_s(z)|&=|P(\phi_s(P^{t-1}_s(z)))-\phi^t_s(z)|\\
&\le |P(\phi_s(P^{t-1}_s(z)))-\phi_s(P^{t-1}_s(z))|+|\phi_s(P^{t-1}_s(z))-\phi^t_s(z)|\\
&\le \epsilon+K|P^{t-1}_s(z)-\phi_s^{t-1}(z)|,
\end{array}
$$
and \eqref{lipschitz_iterates} follows by induction. 

We now exploit these properties to show that for suitable choices of $\beta,\epsilon'$, the previously defined rates $r_{s,z_i}(t)$ verify the following inequalities for all $i$ such that $n(A_i)>0$ and thus $z_i$ is defined:
\begin{equation}\label{rates_6}
\begin{array}{ll}
\sum_{z\in A_i}\lambda_z\le \sum_s r_{s,z_i}(1),\\
\sum_{s}\sum_{z_j:P^1_s(z_j)=z_i}r_{s,z_j}(t)\psi_s(z_j)\le \sum_s r_{s,z_i}(t+1)
\end{array}
\end{equation}
The first equation in \eqref{rates_6} reads, in view of \eqref{rates_2}, \eqref{rates_3}:
$$
\sum_{z\in A_i}\lambda_z\le \sum_{z\in A_i}\sum_s\mu_{s,z}(1),
$$
which holds with equality by \eqref{rates_1} i).

The left-hand side of the second equation in \eqref{rates_6} reads for $t\le T$:
$$
\sum_{s}\sum_{z_j:P^1_s(z_j)=z_i}\sum_{z:P_s^t(z)=z_j}\mu_{s,\phi^t_s(z)}(t)\psi_s(z_j)+\epsilon'(1+\beta^t).
$$
Using the Lipschitz property of $\psi$, the bound \eqref{lipschitz_iterates} established between  $P^t_s(z)$ and $\phi^t_s(z)$, and letting $\Lambda:=\sum_z \lambda_z$, this is no larger than
$$
\sum_{s}\sum_{z_j:P^1_s(z_j)=z_i}\sum_{z:P_s^t(z)=z_j}\mu_{s,\phi^t_s(z)}(t)\psi_s(\phi^t_s(z))+\epsilon'(1+\beta^t)+\Lambda(1-\alpha)^t\epsilon K\frac{K^t-1}{K-1}.
$$
Indeed, the sum $\sum_z \mu_{s,z}(t)$ of rates at step $t$ is at most $\Lambda(1-\alpha)^t$. This last expression can be rearranged to give
$$
\sum_s\sum_{z:P^{t+1}_s(z)=z_i}\mu_{s,\phi^t_s(z)}(t)\psi_s(\phi^t_s(z))+\epsilon'(1+\beta^t)+\Lambda(1-\alpha)^t\epsilon K\frac{K^t-1}{K-1}.
$$
In view of \eqref{rates_1} ii), the first summation is equal to
$$
\sum_s\sum_{z:P^{t+1}_s(z)=z_i}\mu_{s,\phi^{t+1}_s(z)}=\sum_s r_{s,z_i}(t+1)-\epsilon'(1+\beta^{t+1}).
$$
The difference between the right-hand side and the left-hand side of the second equation in \eqref{rates_6} is therefore lower-bounded by
$$
\epsilon'(1+\beta^{t+1})-\epsilon'(1+\beta^t)-\Lambda(1-\alpha)^t\epsilon K\frac{K^t-1}{K-1}=
\epsilon'\beta^t(\beta-1)-\Lambda(1-\alpha)^t\epsilon K\frac{K^t-1}{K-1}.
$$
Assuming $K\ge 2$, $\beta=K+1$, and $\epsilon'=\Lambda \epsilon$, this difference is at least
$$
\Lambda \epsilon (K+1)^t K-\Lambda \epsilon K(K^t-1)\ge K\Lambda \epsilon[(K+1)^t-K^t+1]\ge K\Lambda \epsilon.
$$
Letting $\delta:=K\Lambda \epsilon$, we have in fact shown a strengthening of the second equation in \eqref{rates_6}, namely:
\begin{equation}
\label{rates_66}
t\in{2,\ldots, T}\hbox{ and } n(A_i)>0\Rightarrow \sum_{s}\sum_{z_j:P^1_s(z_j)=z_i}r_{s,z_j}(t)\psi_s(z_j)\le \delta +\sum_s r_{s,z_i}(t+1)
\end{equation}
 
Consider now $t\ge T+1$. The left-hand side of the second equation in \eqref{rates_6} verifies
$$
\sum_{z_j:P^1_{s_0}(z_j)=z_i}r_{s_0,z_j}(t)\psi_s(z_j)\le \sum_{z_j:P^1_{s_0}(z_j)=z_i}r_{s_0,z_j}(t)(1-\alpha),
$$ 
by the lower-bound of $\alpha$ on the $p_{sc}$. This implies that the announced inequality also holds for $t>T$.

We now verify that, provided $\epsilon$ was chosen small enough, the constructed rates $r_{s,z_i}(t)$ satisfy the capacity constraints of the servers. For $s\ne s_0$, this is easily verified, as by \eqref{rates_1} iii),
$$
\sum_t\sum_i r_{s,z_i}(t)\le\sum_z\sum_t \mu_{s,z}(t)\le \mu_s.
$$
Consider now server $s_0$. We then have
$$
\sum_t\sum_i r_{s_0,z_i}(t)\le \sum_{z}\sum_t\mu_{s_0,z}(t)+\sum_{t=1}^{T+1}\epsilon'(1+\beta^t)+\frac{\Lambda(1-\alpha)^{T+1}+\epsilon'(1+\beta^{T+1})}{\alpha}\cdot
$$
Thus by \eqref{rates_1} iv) this meets the capacity constraint of server $s_0$ provided
$$
\sum_{t=1}^{T+1}\epsilon'(1+\beta^t)+\frac{\Lambda(1-\alpha)^{T+1}+\epsilon'(1+\beta^{T+1})}{\alpha}\le R.
$$
This can clearly be achieved by first choosing $T$ such that $\Lambda(1-\alpha)^{T+1}\le R\alpha/2$, and then $\epsilon$ such that 
$$
\sum_{t=1}^{T+1}\epsilon'(1+\beta^t)+\frac{\epsilon'(1+\beta^{T+1})}{\alpha}\le R/2.
$$
It finally remains to prove that the Foster-Lyapunov stability criterion holds for our proposed backpressure policy. Assume that each server $s$ dedicates capacity $\sum_{t\ge 1} r_{s,z_i}(t)$ to jobs of type $z_i$.  This does not exceed servers' capacities as we just showed. Moreover, in view of \eqref{rates_6} and \eqref{rates_66}, under this allocation the drift of any $n(A_i)$ such that $n(A_i)>0$   reads
$$
\sum_{z\in A_i}\lambda_z -\sum_{t\ge 1}\sum_s r_{s,z_i}(t)+\sum_{t\ge 1}\sum_s\sum_{j:P^1_s(z_j)=z_i}r_{s,z_j}(t)\psi_s(z_j)\le -T\delta\1_{n(A_i)>0}.
$$
For an arbitrary policy, let $\mu_i$ denote the service rate it devotes to those $z$ in $A_i$, and $\lambda'_i$ denote the overall arrival rate of jobs with type $z$ in $A_i$ whether from external arrivals or unsuccessful treatments. The drift for our candidate Lyapunov function $L(n)=\sum_i n(A_i)^2$ then reads
$$
\sum_i (\lambda'_i+\mu_i)+2n(A_i)[\lambda'_i-\mu_i]\le \Lambda +2 \sum_s \mu_s + 2\sum_in(A_i)[\lambda'_i-\mu_i],
$$
where we used the fact that the overall arrival rate cannot be larger than the exogeneous arrival rate plus the overall service rate. 

Under the allocations $\sum_{t\ge 1}r_{s,z_i}(t)$ we just considered, the summation in the right-hand side is at most $-2\delta T\sum_i n(A_i)$. Since the backpressure policy we have introduced minimizes this summation among all feasible policies, it guarantees a drift for the Lyapunov function $L$ of at most $\Lambda+2\sum_s\mu_s - 2\delta T\sum_i n(A_i)$. We can therefore rely on Foster's criterion to deduce that the return time to the set $\mathcal A = \{n: \sum_i n(A_i)\le (\Lambda+2\sum_s\mu_s)/(\delta T)\}$ has bounded expectation. We will be done if we show that the system empties infinitely often. For this, we use the argument similar to that used in Theorem~\ref{thm:BP}.

Fix a constant $\beta >0$. Since $\alpha >0$, we have that for any interval of time of size $\beta$ the probability that no arrival happens in this interval and that a task leaves the system is finite. 

Suppose that system is in a state $n \in \mathcal A$ at time $t=0$. Now consider renewal times $T_i, i=0,1, 2, \ldots,$, where $T_0 = 0$ and for each  $i>0$, $T_i$ is defined as follows: $T_i$ is equal to $T_{i-1} + \beta$ if indeed no arrival happens and a task leaves the system in the interval $[T_{i-1}, T_{i-1} + \beta)$, else $T_i$ is the first time of return to $\mathcal A$ after $T_{i-1}$. Clearly $E[T_i] $ since $T$ as defined above is finite. Further probability that a task leaves system in time $T_i - T_{i-1}$ is finite, say $\tilde \alpha$. Thus, time for system emptying after first reaching $\mathcal A$ can be upper-bounded by sum of $K$ geometric random variables with rate $\tilde \alpha$. Thus expected time to return to state $0$ is finite. Hence, the system is stable. 
\qed

\subsection{Proof of Proposition~\ref{prop:OptAsymmetricA}} \label{proof:OptAsymmetricA}

We use Theorem \ref{thm:OptimalStability} to prove this result. We first establish the sufficient condition and then the necessary condition. For Asymmetric($a$) system we have $\mathcal Z = \{z',z''\}$ where $z'_c = \frac{1}{2}$ for each $c\in C$, and $z''_{c} = \indic{c=c_2}$.   The flow conservation constraints in Theorem \ref{thm:OptimalStability} can be given as follows: 

$$ \lambda  = \sum_{s} \nu_{s,z'}, \text{ and } \sum_{s} \nu_{s,z'} \psi_s(z') + \sum_{s} \nu_{s,z''}\psi_{s}(z'')  = \sum_{s} \nu_{s,z''} $$

Suppose $a \ge \frac{1}{2}$. There exists an $\epsilon > 0$ such that $\lambda = \frac{3a(1-\epsilon)}{a+1}$.  It can be checked that $(\nu_{sz})_{s,z}$ where  
$$\nu_{s_2,z'} = 1 - \epsilon , \nu_{s_2,z''} = 0, \nu_{s_1,z'} = \frac{2a -1}{a+1}(1-\epsilon), \nu_{s_1,z''} = \frac{2-a}{a+1}(1-\epsilon)$$
and $(\delta_s)_{s\in S}$ where $\delta_s = \epsilon$ for each $s$ satisfy sufficient conditions of Theorem~\ref{thm:OptimalStability}. 

Now suppose $a< \frac{1}{2}$. There exists an $\epsilon > 0$ such that $\lambda = 2a (1-\epsilon)$.  It can be checked that $(\nu_{s,z})_{s,z}$ where  
$$\nu_{s_2,z'} = 2a (1-\epsilon) , \nu_{s_2,z''} = 0, \nu_{s_1,z'} = 0, \nu_{s_1,z''} = (1-\epsilon)$$
and $(\delta_s)_{s\in S}$ where $\delta_s = \epsilon$ for each $s$ satisfies sufficient conditions of Theorem~\ref{thm:OptimalStability}. 

The sufficient condition then follows from the proof of Theorem~\ref{thm:OptimalStability} by taking $\mathcal Y$ as $\mathcal Z$.

We now show the necessary condition. From the necessary conditions in  Theorem \ref{thm:OptimalStability}, we have the following:

\begin{align}
\lambda & = \sum_{s} \nu_{s,z'}, \label{eq:FlowConv1} \\ 
\frac{(1-a)}{2} \nu_{s_1,z'} + \frac{1}{2} \nu_{s_2,z'} + (1-a) \nu_{s_1,z''}& = \nu_{s_1,z''} , \label{eq:FlowConv2}  \\
\nu_{s_1,z'} + \nu_{s_1,z''} & \le 1, \label{eq:cap1} \\
\nu_{s_2,z'} + \nu_{s_2,z''}  & \le 1. \label{eq:cap2} 
\end{align}

From \eqn{eq:FlowConv2} we get:
$$ \nu_{s_1,z''} = \frac{1}{a} \left( \frac{(1-a)}{2} \nu_{s_1,z'}  +  \frac{1}{2} \nu_{s_2,z'}  \right).$$

By substituting in \eqn{eq:cap1} this the above expression for $\nu_{s_1,z''} $, we get

$$ \nu_{s_1,z'} + \frac{1}{a} \left( \frac{(1-a)}{2} \nu_{s_1,z'}  +  \frac{1}{2} \nu_{s_2,z'} \right) \le 1   $$

Upon simplifying, we get

\begin{equation}\label{eq:nus1zBound}
 \nu_{s_1,z'}  \le  \frac{2a}{1+a} \left(1 - \frac{1}{2a} \nu_{s_2,z'}  \right). 
 \end{equation}

Further, we need $\nu_{s_1,z'} $ to be non-negative. Thus, we need $\nu_{s_2,z'} \le 2a$. 

Substituting \eqn{eq:nus1zBound} in \eqn{eq:FlowConv1} we get

$$\lambda \le  \max\left( \frac{2a}{1+a} \left(1 - \frac{1}{2a} \nu_{s_2,z'} \right),0\right) + \nu_{s_2,z'}  .$$

Suppose $a \ge \frac{1}{2}$. Then, subject to \eqn{eq:cap2} and $\nu_{s_2,z'} \le 2a$, the right hand side of the above is maximized when $ \nu_{s_2,z'} = 1$ and $ \nu_{s_2,z''} = 0$. We thus obtain $\lambda \le \frac{3a}{a+1}$.  Similarly, if $a < \frac{1}{2}$, then  subject to \eqn{eq:cap2} and $\nu_{s_2,z'} \le 2a$, the right hand side of the above is maximized when $ \nu_{s_2,z'} = 2a$ and $  \nu_{s_2,z''} = 0$, from which we obtain $\lambda \le 2a$. Thus, overall, we have $\lambda \le \min(\frac{3a}{a+1}, 2a)$.  Hence the result follows.
\qed

\subsection{Proof of Proposition~\ref{prop:Random}} \label{proof:Random}
We show the result for a general a task-expert system. The result for Asymmetric($a$) system then follows immediately.

Note that the system under random policy is equivalent to the one where pure-type of a task is revealed upon arrival, \ie, there is no uncertainty in task types. This is true since the random policy does not use the information of type (pure or mixed). We thus let the pure-type of each task be revealed upon arrival. Let $X_c(t)$ be the number of tasks in the system of pure-type $c$.  Let $X(t) = (X_c(t))_c$. For each $c\in C$, the arrival rate into queue $X_c(t)$ is equal to 
 $$\lambda_c \define \sum_{z\in Z} \lambda z_c \pi_z.$$
 
We first show the if part of the result.  Suppose that we have $ \sum_{c\in C} \frac{\lambda_c}{\sum_{s\in S} \mu_s p_{s,c}} < 1$. We use the fluid limit approach developed in \cite{RyS92,Dai95,Mas07}. Roughly, given initial condition $X(0) = x$, the fluid trajectories of the state process $X(t)$ can be obtained by scaling initial conditions, speeding time, and then studying the rescaled process; \ie, letting $\lim_{k\to \infty} \frac{1}{k} X(0) = x$, and studying $\lim_{k\to \infty}  \frac{1}{k} X(kt)$. 
   
Using arguments similar to those used in \cite{Mas07}, the fluid limits for the number of tasks in each class can be shown to satisfy the following at almost all times $t$: 
for each $c \in C$ and $X_c > 0$ we have
\begin{equation}\label{eq:Xdot}
\frac{d}{dt} X_c = \lambda_c - \sum_{s\in S} \mu_s p_{s,c} \frac{X_c}{\sum_{c'}X_{c'}}.
\end{equation}

Define a function $L$ on $\mathbb R^C$ as
\begin{equation}\label{eq:LX}
L(X) = \sum_c X_c \log \left(\frac{X_c}{\gamma_c\sum_{c'}X_{c'}} \right),
\end{equation}
where $\gamma_c \define  \frac{\lambda_c}{\sum_{s\in S} \mu_s p_{s,c}}$. 

Further, by following the arguments similar to \cite{Mas07}, if we have that $L(X) \to \infty$ and $\frac{d}{dt} L(X) \le -\epsilon $  for all $X$ such that $|X| = 1$ under fluid limits then the stability of the original system follows.  We show below that both these limits hold.

Using \eqn{eq:Xdot} and \eqn{eq:LX}, we obtain
\begin{align}
\frac{d}{dt} L(X) & = \sum_c \left(\frac{d}{dt} X_c\right)  \log \left(\frac{X_c}{\gamma_c\sum_{c'}X_{c'}} \right), \;\;\;\; \\
       			 & =\sum_c \left(  \lambda_c - \sum_{s\in S} \mu_s p_{s,c} \frac{X_c}{\sum_{c'}X_{c'}} \right) \left( \log \frac{X_c}{\sum_{c'}X_{c'}} - \log \gamma_c \right),\\
   		         &= \sum_c \left(\sum_s \mu_s p_{s,c}\right) \left(  \frac{\lambda_c}{\sum_{s\in S} \mu_s p_{s,c}} -  \frac{X_c}{\sum_{c'}X_{c'}} \right) \left( \log \frac{X_c}{\sum_{c'}X_{c'}} - \log \gamma_c \right),  \\ 		   	         &= \sum_c \left(\sum_s \mu_s p_{s,c}\right) \left(  \gamma_c -  Y_c \right)  \log (Y_c / \gamma_c ), \label{eq::8_11}
\end{align}

where $Y_c := \frac{X_c}{\sum_{c'}X_{c'}}$. Now, \eqref{eq::8_11} is negative and strictly bounded away from zero. This can be seen as follows. Firstly, all terms in the sum are non-positive. Therefore, it suffices to show that there exists a $\delta > 0$ such that there always exists a $c$ for which $\left(  \gamma_c -  Y_c \right)  \log( Y_c / \gamma_c  ) \leq - \delta.$ Since, $\sum_c Y_c = 1$ and, for some fixed $\epsilon > 0$, $\sum_c \gamma_c = 1 - \epsilon$, it follows that there exists $c$ such that $\gamma_c - Y_c \leq - \epsilon.$ For this $c$, we thus also have $Y_c / \gamma_c \geq 1 + \epsilon$.  Consequently, $\left(  \gamma_c -  Y_c \right)  \log( Y_c / \gamma_c  ) \leq -  \epsilon \log(1+ \epsilon).$  

Let $\theta = 1/\sum_c \gamma_c$ and $\hat{\gamma}_c = \theta \gamma_c$ for each $c\in C$. Since $\sum_c \gamma_c < 1$, we have $\theta > 1$. Let $D(p||q)$ be the Kullback-Leibler divergence between two Bernoulli distributions with parameters $p$ and $q$, \ie, $D(p||q) = p \log (\frac{p}{q}) + (1-p) \log(\frac{1-p}{1-q})$.  Now, we can write

\begin{align}
L(X)  & = \sum_c X_c \log \left(\frac{\theta X_c}{\hat \gamma_c\sum_{c'}X_{c'}}\right) \\
	 &= \sum_c X_c \log \theta + \sum_c X_c \log \left(\frac{X_c}{\sum_{c'}X_{c'}} . \frac{1}{\hat \gamma_c}\right) \\
	&= \sum_c X_c \log \theta +  \left(\sum_c X_c\right) D\left(\left(\frac{X_c}{\sum_{c'}X_{c'}}\right)_{c\in C} \Bigg|\Bigg| (\hat \gamma_c)_{c\in C}\right)
\end{align}
which converges to $\infty$ as $|X|$ grows large.

Hence, the if part of the result follows.

We now show that the system is unstable if $\sum_{c\in C} \frac{\lambda_c}{\sum_{s\in S} \mu_s p_{s,c}} \ge 1$. 
We consider the original system instead of the fluid limits. Consider the following function:

$$K(X) = \sum_{c} \frac{1}{\sum_s \mu_s p_{s,c}} X_c.$$

Clearly, $K(X) \to \infty$ as $X \to \infty$. Define $D(.)$ as in \eqn{eqn:DriftOfL}, but for $K$ instead of $L$.  Then, we have

$$ \frac{1}{\tau_{\bar 0}}D(\bar 0) = \sum_c  \frac{ \lambda_c}{\sum_s \mu_s p_{s,c}},$$

and for $X \neq \bar 0$, we have 

\begin{align*}
 \frac{1}{\tau_X} D(X) & =  \sum_{c}  \frac{1}{\sum_s \mu_s p_{s,c}} \left( \lambda_c - \sum_{s\in S} \mu_s p_{s,c} \frac{X_c}{\sum_{c'}X_{c'}} \right), \\
   & =  \sum_{c}    \frac{ \lambda_c}{\sum_s \mu_s p_{s,c}} - \sum_c \frac{X_c}{\sum_{c'}X_{c'}} \\
   & = \sum_c  \frac{ \lambda_c}{\sum_s \mu_s p_{s,c}} - 1, \\
  & \ge 0.
\end{align*}

Thus, the drift is non-negative for all but finite number of states.  Further, since $K(X)$ is bounded from below, the maximum change in $K(X)$ upon an arrival or a departure is also bounded, using Proposition I.5.4 on page 22 in \cite{asm03}, we get the only if part.  \qed

\subsection{Proof of Proposition~\ref{prop:PreemptiveGreedy}} \label{proof:PreemptiveGreedy}

We first show the if part. For each $t$ let $t+\tau(t)$ be the time at which the first event (arrival or completion of a response) occurs after time $t$.  Let $\tau_n = E[\tau(t)| N(t) = n]$, \ie, given that $N(t) = n$ at time $t$, $\tau_n$  is the expected time at which the first event occurs after time $t$. 
For example, for $n = 0$ we have $\tau_n = 1/\lambda$. 


Now suppose that $\lambda < 4a/(2+a)$. Then, it can be checked that $\frac{2-a}{2(2-\lambda)} \lambda < a$. Thus, there exists $\delta>0$ such $(1+\delta) \frac{2-a}{2(2-\lambda)} \lambda < a$.
Now, consider the following candidate Lyapunov function: for each $n$, we have
$$
\frac{1}{\tau_n} L(n) = (1+\delta)\frac{2-a}{2(2-\lambda)} n_{z'}  + n_{z''},
$$
where $\delta$ is a constant obtained as above. 

Let 
\begin{equation}\label{eqn:DriftOfL}
D(n) =  E\left[L( N(t+\tau(t)))  - L( N(t))\big|  N(t) =  n \right]
\end{equation}

 Consider the states $n$ such that $n_{z'} > 0$. For these states, we obtain 

\begin{align*}
\frac{1}{\tau_n} D(n) & =  (1+\delta) \frac{2-a}{2(2-\lambda)} \left(\lambda - \mu_{s_1} - \mu_{s_2} \right) 
+  \left( \mu_{s_1}\psi_{s_1}(z')  + \mu_{s_2}\psi_{s_2}(z') \right) \\
&  = (1+\delta) \frac{2-a}{2(2-\lambda)}(\lambda -2) +  \frac{1-a}{2} + \frac{1}{2}  
= -\delta \frac{2-a}{2} 
< 0.
\end{align*}

Now, consider states $n$ such that $n_{z'} = 0$ and $n_{z''}> 0$. For these states we have
\begin{align*}
\frac{1}{\tau_n} D(n) & = (1+ \delta) \frac{2-a}{2(2-\lambda)} \lambda - \mu_{s_1}a  < 0.
\end{align*}

Since the drift outside of the state $(0,0)$ is less than or equal to $-\min( \delta (2-a)/2), \mu_{s_1}a - (1+ \delta) \frac{2-a}{2(2-\lambda)} \lambda ) < 0$, from the Lyapunov-Foster Theorem we obtain that $N(t)$ is positive recurrent if $\lambda < 4a/(2+a)$. 


We now show the only if part. Suppose that $\lambda \ge 4a/(2+a)$. Then, there exists $\delta\le 0$ such $(1+\delta) \frac{2-a}{2(2-\lambda)} \lambda \ge a$. Thus, drift is non-negative for all but finite values of $n$. Further, since $L(\cdot)$ is bounded from below, and since the maximum change in $L(\cdot)$ upon an arrival or a departure is bounded, using Proposition I.5.4 on page 22 in \cite{asm03}, we establish the only if part.

\revised{

\subsection{Proof of Proposition~\ref{prop:NonPreemptiveGreedy}} \label{proof:NonPreemptiveGreedy}

First, consider the following lemma.

\begin{lemma}
Under Non-preemptive Greedy policy, the fraction of time server $s_1$ spends in serving tasks of true type $c_1$ is bounded from below by $\frac{1}{16} \frac{\lambda(2+\lambda)}{1+\lambda} $.
\end{lemma}

Thus, the maximum capacity available to serve tasks of true type $c_2$ is $1-\frac{1}{16} \frac{\lambda(2+\lambda)}{1+\lambda} $. In turn, the system is unstable if $\frac{\lambda}{2} > a\left(1-\frac{1}{16} \frac{\lambda(2+\lambda)}{1+\lambda} \right)$. From this the result follows via some simplifications. We prove the lemma below. 

In what follows we assume that queue $z''$ is saturated, since forcing $z''$ to be saturated only reduces the time the server $s_1$ spends on queue $z'$ under Non-preemptive Greedy policy. Further, note that the quantity of interest is twice the fraction of time server $s_1$ spends in serving tasks of true type $c_1$.  To obtain a bound on this quantity, we separately obtain an upper bound on the expected length of a busy-idle cycle at queue $z'$ and a lower bound  bound on the expected time $s_1$ spends in serving queue $z'$ within a busy-idle cycle, and use the renewal reward theorem. 

The expected length of a busy-idle cycle at queue $z'$ can be bounded from above by that of an alternate system in which server $s_2$ is forced to stay idle while server $s_1$ is serving queue $z''$. In this modified system, the idle time for queue $z'$ is upper bounded by $\frac{1}{\lambda} +1$ since the inter-arrival times are Exponential($\lambda$) and the time server $s_2$ is forced to stay idle is Exponential($1$). Further, the number of arrivals into queue $z'$ in an Exponential($1$) time period is Geometric($\lambda$) distributed. Thus, at the end of idle period, when both the servers start serving the queue, the backlog in the queue is $1+Y$ where $Y$ is Geometric($\lambda$) distributed. In turn, the expected busy period for queue $z'$ is bounded from above by $\frac{1+\lambda}{2-\lambda}$. 

Thus, the expected length of a busy-idle cycle at queue $z'$ is bounded from above by $\frac{1}{\lambda} +1 + \frac{1+\lambda}{2-\lambda}$. 

We now provide a lower bound on the expected time server $s_1$ spends on serving queue $z'$ within its busy-idle cycle. At the beginning of the busy period of queue $z'$, with probability $1$ the server $s_2$ is serving queue $z'$ and server $s_1$ is serving queue $z''$. The time it takes for one of the servers to complete the current service is Exponential($2$) distributed. Let $Y'$ be the number of arrivals into queue in this period. Then, $Y'$ is Geometric($\lambda/2$) distributed. With probability half, server $s_1$ is the one who completed first, and the two servers are now working to drain a backlog of $1+Y$. The average duration for
this to terminate is $\frac{1+\lambda/2}{2-\lambda}$. Thus, $\frac{1+\lambda/2}{2(2-\lambda)}$ is a lower bound on the expected time server $s_1$ spends on serving queue $z'$ within its busy-idle cycle.

Thus, the fraction of time $s_1$ spends in serving queue $z'$ is $\frac{\frac{1+\lambda/2}{2(2-\lambda)}}{\frac{1}{\lambda} +1 + \frac{1+\lambda}{2-\lambda}}$. From this the lemma follows upon some simplifications, and thus the proposition follows as well.  

Some details of simplifications:

To obtain the fraction of time...

$$\frac{\frac{1}{2} \frac{1+\lambda/2}{2(2-\lambda)}}{\frac{1}{\lambda} +1 + \frac{1+\lambda}{2-\lambda}} =\frac{  \frac{1}{8} \frac{2+\lambda}{(2-\lambda)}}{\frac{2-\lambda + \lambda(2-\lambda) + \lambda(1+\lambda) }{\lambda(2-\lambda)}} = \frac{1}{8} \frac{\lambda(2+\lambda)}{2-\lambda + \lambda(2-\lambda) + \lambda(1+\lambda) } =  \frac{1}{8} \frac{\lambda(2+\lambda)}{2+2\lambda} =   \frac{1}{16} \frac{\lambda(2+\lambda)}{1+\lambda} $$

To obtain stability conditions

$$\frac{\lambda}{2} > a\left(1-\frac{1}{16} \frac{\lambda(2+\lambda)}{1+\lambda} \right) \iff \frac{\lambda}{2a} -1 + \frac{1}{16} \frac{\lambda(2+\lambda)}{1+\lambda} > 0 \iff 8a^{-1} \lambda(1+\lambda) - 16(1+\lambda) + \lambda(2+\lambda) > 0 $$
$$\iff  \lambda^2(8a^{-1} + 1) +  \lambda (8a^{-1}  - 16 + 2) -16 > 0  \iff \lambda^2(8a^{-1} + 1) +  \lambda (8a^{-1}  - 14) -16 > 0  $$
}

\bibliographystyle{plain}
\bibliography{mybib}

\begin{thebibliography}{10}

\bibitem{AgG12}
Shipra Agrawal and Navin Goyal.
\newblock Analysis of thompson sampling for the multi-armed bandit problem.
\newblock In {\em Proceedings of the 25th Conference on Learning Theory}, 2012.

\bibitem{ASK12}
M.~Alresaini, M.~Sathiamoorthy, B.~Krishnamachari, and M.J. Neely.
\newblock Backpressure with adaptive redundancy (bwar).
\newblock In {\em Proc.\ IEEE INFOCOM}, pages 2300--2308, March 2012.

\bibitem{asm03}
S{\o}ren Asmussen.
\newblock {\em Applied probability and queues}.
\newblock Springer Science \& Business Media, 2nd edition, 2003.

\bibitem{ACF02}
Peter Auer, Nicol{\`o} Cesa-Bianchi, and Paul Fischer.
\newblock Finite-time analysis of the multiarmed bandit problem.
\newblock {\em Machine Learning}, 47(2):235--256, 2002.

\bibitem{BeW01}
S.~L. Bell and R.~J. Williams.
\newblock Dynamic scheduling of a system with two parallel servers in heavy
  traffic with resource pooling: asymptotic optimality of a threshold policy.
\newblock {\em Ann. Appl. Probab.}, 11(3):608--649, 08 2001.

\bibitem{BiM15}
Kostas Bimpikis and Mihalis~G Markakis.
\newblock Learning and hierarchies in service systems.
\newblock {\em Unpublished manuscript}, 2015.

\bibitem{Bre13}
Pierre Br{\'e}maud.
\newblock {\em Markov chains: Gibbs fields, Monte Carlo simulation, and
  queues}, volume~31.
\newblock Springer Science \& Business Media, 2013.

\bibitem{Buc12}
S{\'e}bastien Bubeck and Nicol{\`o} Cesa-Bianchi.
\newblock Regret analysis of stochastic and nonstochastic multi-armed bandit
  problems.
\newblock {\em Foundations and Trends in Machine Learning}, 5(1), 2012.

\bibitem{BSS11}
L.X. Bui, R.~Srikant, and A.~Stolyar.
\newblock A novel architecture for reduction of delay and queueing structure
  complexity in the back-pressure algorithm.
\newblock {\em IEEE/ACM Transactions on Networking}, 19(6):1597--1609, Dec
  2011.

\bibitem{ChK13}
Yuxin Chen and Andreas Krause.
\newblock Near-optimal batch mode active learning and adaptive submodular
  optimization.
\newblock In {\em ICML'13}, pages 160--168, 2013.

\bibitem{CYL15}
Y.~Cui, E.M. Yeh, and R.~Liu.
\newblock Enhancing the delay performance of dynamic backpressure algorithms.
\newblock {\em IEEE/ACM Transactions on Networking}, 2015.

\bibitem{Dai95}
J.~G. Dai.
\newblock On positive harris recurrence of multiclass queueing networks: A
  unified approach via fluid limit models.
\newblock {\em The Annals of Applied Probability}, 5(1):49--77, 1995.

\bibitem{DDP15}
M~Daltayanni, L~De~Alfaro, and P~Papadimitriou.
\newblock Workerrank: Using employer implicit judgements to infer worker
  reputation.
\newblock In {\em WSDM}, 2015.

\bibitem{FuK16}
Kaito Fujii and Hisashi Kashima.
\newblock Budgeted stream-based active learning via adaptive submodular
  maximization.
\newblock In {\em NIPS'16}, pages 514--522, 2016.

\bibitem{Gao16}
Chao Gao, Yu~Lu, and Dengyong Zhou.
\newblock Exact exponent in optimal rates for crowdsourcing.
\newblock In {\em ICML'16}, pages 603--611, 2016.

\bibitem{GNT06}
Leonidas Georgiadis, Michael~J. Neely, and Leandros Tassiulas.
\newblock Resource allocation and cross-layer control in wireless networks.
\newblock {\em Foundations and Trends in Networking}, 1(1):1--144, 2006.

\bibitem{GGW11}
John Gittins, Kevin Glazebrook, and Richard Weber.
\newblock {\em Multi-armed bandit allocation indices}.
\newblock John Wiley \& Sons, 2011.

\bibitem{GKR10}
Daniel Golovin, Andreas Krause, and Debajyoti Ray.
\newblock Near-optimal bayesian active learning with noisy observations.
\newblock In {\em NIPS'10}, pages 766--774, 2010.

\bibitem{Har98}
J.~Michael Harrison.
\newblock Heavy traffic analysis of a system with parallel servers: asymptotic
  optimality of discrete-review policies.
\newblock {\em Ann. Appl. Probab.}, 8(3):822--848, 08 1998.

\bibitem{Ho12}
Chien-Ju Ho and Jennifer~Wortman Vaughan.
\newblock Online task assignment in crowdsourcing markets.
\newblock In {\em AAAI'12}, pages 45--51, 2012.

\bibitem{JCK14}
Shervin Javdani, Yuxin Chen, Amin Karbasi, Andreas Krause, Drew Bagnell, and
  Siddhartha~S Srinivasa.
\newblock Near optimal bayesian active learning for decision making.
\newblock In {\em AISTATS'14}, pages 430--438, 2014.

\bibitem{JJS13}
Bo~Ji, Changhee Joo, and N.B. Shroff.
\newblock Delay-based back-pressure scheduling in multihop wireless networks.
\newblock {\em IEEE/ACM Transactions on Networking}, 21(5):1539--1552, Oct
  2013.

\bibitem{JKK17}
Ramesh Johari, Vijay Kamble, and Yash Kanoria.
\newblock Matching while learning.
\newblock {\em CoRR}, abs/1603.04549, 2017.

\bibitem{KOS14}
David~R. Karger, Sewoong Oh, and Devavrat Shah.
\newblock Budget-optimal task allocation for reliable crowdsourcing systems.
\newblock {\em Operations Research}, 62(1):1--24, 2014.

\bibitem{KlS03}
Jon Kleinberg and Mark Sandler.
\newblock Convergent algorithms for collaborative filtering.
\newblock In {\em Proceedings of the 4th ACM Conference on Electronic
  Commerce}, 2003.

\bibitem{Kls04}
Jon Kleinberg and Mark Sandler.
\newblock Using mixture models for collaborative filtering.
\newblock In {\em Proceedings of the 36th annual ACM Symposium on Theory of
  Computing}, 2004.

\bibitem{Kum14}
Anurag Kumar.
\newblock Discrete event stochastic processes: Lecture notes for an engineering
  curriculum.
\newblock 2014.

\bibitem{LaR85}
T.L Lai and Herbert Robbins.
\newblock Asymptotically efficient adaptive allocation rules.
\newblock {\em Advances in Applied Mathematics}, 6(1):4 -- 22, 1985.

\bibitem{MBS16}
Siva~Theja Maguluri, Sai~Kiran Burle, and R.~Srikant.
\newblock Optimal heavy-traffic queue length scaling in an incompletely
  saturated switch.
\newblock {\em SIGMETRICS Perform. Eval. Rev.}, 44(1):13--24, June 2016.

\bibitem{MaS04}
Avishai Mandelbaum and Alexander~L. Stolyar.
\newblock Scheduling flexible servers with convex delay costs: Heavy-traffic
  optimality of the generalized c$\mu$-rule.
\newblock {\em Operations Research}, 52(6):836--855, 2004.

\bibitem{Mas07}
Laurent Massouli{\'e}.
\newblock Structural properties of proportional fairness: Stability and
  insensitivity.
\newblock {\em Annals of Applied Probability}, 17(3):809--839, 2007.

\bibitem{MaX16}
Laurent Massouli{\'e} and Kuang Xu.
\newblock {On the capacity of information processing systems}.
\newblock In {\em {29th Annual Conference on Learning Theory}}, New York,
  United States, June 2016.

\bibitem{MeP12}
A.~Mehta and D.~Panigrahi.
\newblock Online matching with stochastic rewards.
\newblock In {\em IEEE 53rd Annual Symposium on Foundations of Computer
  Science}, pages 728--737, Oct 2012.

\bibitem{MWZ15}
Aranyak Mehta, Bo~Waggoner, and Morteza Zadimoghaddam.
\newblock {\em Online Stochastic Matching with Unequal Probabilities}, pages
  1388--1404.
\newblock 2015.

\bibitem{NMR03}
M.~J. Neely, E.~Modiano, and C.~E. Rohrs.
\newblock Dynamic power allocation and routing for time varying wireless
  networks.
\newblock In {\em Proc. IEEE INFOCOM}, 2003.

\bibitem{NeW88}
George~L Nemhauser and Laurence~A Wolsey.
\newblock {\em Integer and combinatorial optimization}, volume~18.
\newblock Wiley, 1988.

\bibitem{Rob03}
Philippe Robert.
\newblock Stochastic networks and queues, stochastic modelling and applied
  probability series, vol. 52, 2003.

\bibitem{RyS92}
Aleksandr~Nikolaevich Rybko and Alexander~L Stolyar.
\newblock Ergodicity of stochastic processes describing the operation of open
  queueing networks.
\newblock {\em Problemy Peredachi Informatsii}, 28(3):3--26, 1992.

\bibitem{Shah15}
Nihar~B. Shah, Dengyong Zhou, and Yuval Peres.
\newblock Approval voting and incentives in crowdsourcing.
\newblock In {\em ICML'15}, pages 10--19, 2015.

\bibitem{SVK16}
Virag Shah, Gustavo de~Veciana, and George Kesidis.
\newblock A stable approach for routing queries in unstructured p2p networks.
\newblock {\em IEEE/ACM Transactions on Networking}, 24(5):3136--3147, Oct
  2016.

\bibitem{SHG09}
David~H. Stern, Ralf Herbrich, and Thore Graepel.
\newblock Matchbox: Large scale online bayesian recommendations.
\newblock In {\em Proceedings of the 18th International Conference on World
  Wide Web}, 2009.

\bibitem{TaE92}
L.~Tassiulas and A.~Ephremides.
\newblock Stability properties of constrained queueing systems and scheduling
  policies for maximum throughput in multihop radio networks.
\newblock {\em IEEE Transactions on Automatic Control}, 37:1936--1948, 1992.

\bibitem{TeD10}
Tolga Tezcan and J.~G. Dai.
\newblock Dynamic control of n-systems with many servers: Asymptotic optimality
  of a static priority policy in heavy traffic.
\newblock {\em Operations Research}, 58(1):94--110, 2010.

\bibitem{HJX05}
Heng-Qing Ye, Jihong Ou, and Xue-Ming Yuan.
\newblock Stability of data networks: Stationary and bursty models.
\newblock {\em Operations Research}, 53(1):107--125, 2005.

\bibitem{YSR11}
Lei Ying, Sanjay Shakkottai, Aneesh Reddy, and Shihuan Liu.
\newblock On combining shortest-path and back-pressure routing over multihop
  wireless networks.
\newblock {\em IEEE/ACM Transactions on Networking}, 19(3):841--854, June 2011.

\bibitem{ZCZ16}
Yuchen Zhang, Xi~Chen, Dengyong Zhou, and Michael~I. Jordan.
\newblock Spectral methods meet em: A provably optimal algorithm for
  crowdsourcing.
\newblock {\em J. Mach. Learn. Res.}, 17(1):3537--3580, January 2016.

\end{thebibliography}


\end{document}